\documentclass{article}

\usepackage{arxiv}

\usepackage[utf8]{inputenc} 
\usepackage[T1]{fontenc}    
\usepackage{hyperref}       
\usepackage{url}            
\usepackage{booktabs}       
\usepackage{amsfonts}       
\usepackage{nicefrac}       
\usepackage{microtype}      
\usepackage{xcolor}         
\usepackage{amsmath}
\usepackage{amssymb}
\usepackage{graphicx}
\usepackage{multirow}
\usepackage{colortbl}
\usepackage{pifont}
\usepackage{caption}
\usepackage{float}
\usepackage{algorithm}
\usepackage{algorithmic}
\usepackage{wrapfig}
\usepackage{enumitem}
\definecolor{best}{RGB}{255,228,181}
\definecolor{second}{RGB}{255,243,224}
\usepackage{placeins}
\usepackage{amsthm}
\usepackage[square,numbers,compress]{natbib}

\title{IO-SVD: Input-Output Whitened SVD for Adaptive-Rank LLM Compression}

\author{Ali Abbasi\\
Vanderbilt University\\
\texttt{ali.abbasi@vanderbilt.edu}
\And
Chayne Thrash$^\ast$\\
Vanderbilt University\\
\texttt{chayne.thrash@vanderbilt.edu}
\And
Haoran Qin$^\ast$\\
Vanderbilt University\\
\texttt{haoran.qin@vanderbilt.edu}
\And
Hamed Pirsiavash\\
University of California, Davis\\
\texttt{hpirsiav@ucdavis.edu}
\And
Soheil Kolouri\\
Vanderbilt University\\
\texttt{soheil.kolouri@vanderbilt.edu}
}

\date{}

\begin{document}

\maketitle
\let\thefootnote\relax\footnotetext{$^\ast$Equal contribution.}

\begin{abstract}
Large language models deliver strong performance across language and reasoning tasks, but their storage and compute costs remain major barriers to deployment in resource-constrained and latency-sensitive settings. SVD-based post-training compression offers a hardware-agnostic way to reduce model size and improve inference efficiency through low-rank factorization. However, existing methods often rely on input-only whitening spaces, homogeneous rank allocation, or loss-agnostic allocation heuristics, limiting their ability to preserve model quality under aggressive compression. We propose Input-Output Whitened SVD (IO-SVD), a post-training compression method that forms a KL-aware double-sided whitening space for model weights. Using a second-order expansion of the KL loss over the top-$K$ token probabilities, IO-SVD constructs an output-side metric that captures predictive sensitivity, while input whitening captures activation statistics. We further introduce an efficient heterogeneous rank-allocation strategy that scores whitened singular components using first-order calibration loss estimates and prunes the least sensitive components under a global budget. Inspired by prior work that combines SVD truncation with quantization, we improve hybrid SVD-quantization compression through loss-aware remapping, which selects low-rank factor rows for 8-bit quantization based on the predicted loss change incurred by quantizing them. Extensive experiments across diverse LLM and VLM families, and inference-time analysis shows that IO-SVD compresses LLMs with minimal performance degradation while delivering practical inference speedups. Code is available at \url{https://github.com/mint-vu/IO-SVD.git}
\end{abstract}

\section{Introduction}
\vspace{-.1in}

Large language models (LLMs) have achieved strong performance across natural language understanding, reasoning, and code generation, but their storage and computational costs limit deployment in latency-sensitive settings such as robotics, edge computing, and interactive systems. These constraints motivate post-training compression methods that improve inference efficiency without costly retraining or changes to pretraining pipelines.


Existing compression methods include quantization, pruning, and knowledge distillation. Quantization is effective in practice \cite{llm_int8, gptq, smooth_quant, quip, quip_2, shaoomniquant}, but its efficiency often depends on specialized hardware support and optimized low-bit kernels \cite{park2022lut, zhao2024atom, lin2405qserve, frantar2025marlin}. Pruning removes parameters based on importance metrics \cite{OBC_2022, sparsegpt, llmpruner, sun2024a}, but can struggle to match other methods at high compression ratios \cite{cheng2024survey}. Knowledge distillation transfers knowledge from a large teacher to a smaller student \cite{hinton2015distilling, gu2024minillm, agarwal2024policy}, but typically requires costly gradient-based training and may degrade under large teacher--student architectural or tokenizer gaps \cite{rao2023parameter, nguyen2026ctpd, shinovercoming}. These limitations motivate post-training alternatives that are broadly deployable and require no retraining.


SVD-based compression offers a post-training alternative to quantization for LLMs \cite{fwsvd, asvd, svdllm, svd_llm_V2, dobisvd, gfwsvd, zerosumsvd, li2026optimal}. By replacing dense weight matrices with low-rank factors, it can reduce floating-point operations, while remaining hardware-agnostic and avoiding specialized kernels. Prior methods have progressed from weight-space or layer-output reconstruction \cite{fwsvd} to input-aware whitening based on activation statistics \cite{asvd, svdllm, svd_llm_V2, dobisvd}. However, these objectives remain local: they do not directly model how layer perturbations affect the model's predictive distribution. Recent and concurrent works on double-sided whitening incorporate both input and output statistics \cite{gfwsvd, li2026optimal}; we build on this direction by deriving the output-side metric from a KL divergence to the uncompressed model's predictive distribution.


In addition to the choice of truncation space, rank allocation constitutes an orthogonal design dimension in SVD-based compression. Several approaches adopt simple or predefined allocation strategies across layers \cite{svdllm, fwsvd}, without fully accounting for the heterogeneous sensitivity of different components. This can limit the effectiveness of compression, particularly at high compression ratios where careful distribution of capacity becomes critical. More recently, Dobi-SVD~\cite{dobisvd} proposes an explicit optimization framework to determine the optimal truncation point for each linear layer. While effective, this approach relies on gradient-based optimization and can be computationally expensive and time-consuming to solve in practice. As a result, there remains a need for efficient and scalable surrogates of layer sensitivity that can guide rank allocation without incurring significant computational overhead.

Pure low-rank compression becomes increasingly lossy at aggressive compression ratios because it must discard much of the singular spectrum. Dobi-SVD mitigates this issue by combining truncation with quantization through a remapping strategy \cite{dobisvd}. However, its heuristic selection of components for quantization is agnostic to quantization-induced error and its impact on the task objective. We therefore introduce a loss-aware remapping strategy for hybrid SVD compression.


We make three contributions:
\vspace{-.1in}
\begin{itemize}[itemsep=0pt]
    \item \textbf{KL-aware input-output whitening.}
    We derive a double-sided whitening objective from a second-order KL approximation, capturing both input activation geometry and output-side predictive sensitivity.
\item {\bf Loss-aware adaptive compression.} We introduce an efficient heterogeneous rank-allocation strategy that greedily removes whitened singular components with the smallest predicted calibration-loss impact under a global budget, and extend this loss-aware criterion to hybrid SVD–quantization by selecting low-rank factor rows for 8-bit quantization based on predicted quantization-induced loss.


    \item \textbf{Extensive evaluation across models and hardware.}
   We validate our method across a diverse set of LLM and VLM families, covering model sizes up to 13B parameters, with inference-time analysis demonstrating effectiveness beyond text-only models and practical efficiency across deployment settings.

\end{itemize}

\vspace{-.15in}
\section{Related Work}
\vspace{-.1in}

\noindent\textbf{Compression of Large Language Models.}
The increasing scale of large language models (LLMs) has substantially improved performance, but also creates significant memory and computational costs that complicate deployment. This has motivated a broad class of post-training compression methods that reduce model size without full retraining, including pruning, quantization, and low-rank factorization. Optimal Brain Compression (OBC)~\citep{OBC_2022} performs greedy post-hoc pruning and quantization by minimizing a second-order reconstruction objective on a small calibration set, but does not scale directly to modern LLMs due to the cost of repeated inverse-Hessian computations. SparseGPT~\citep{sparsegpt} adapts this idea to very large models using blockwise approximations, while LLM-Pruner~\citep{llmpruner} performs structured pruning based on neuron connectivity and Wanda~\citep{sun2024a} proposes a one-shot pruning criterion combining weight magnitude and activation statistics. Quantization methods similarly rely on calibration data and approximate second-order or activation-aware objectives: LLM.int8()~\citep{llm_int8} uses mixed precision to handle activation outliers, SmoothQuant~\citep{smooth_quant} rescales weights and activations to mitigate outliers, GPTQ/OPTQ~\citep{gptq} adapts OBC using row-wise updates and block Hessian approximations, and QuIP~\citep{quip, quip_2} improves quantization through incoherence processing. Although pruning and quantization can achieve strong compression, their realized speedups often depend on hardware support and specialized kernels. Low-rank compression offers a complementary path because replacing dense linear layers with products of smaller matrices can directly reduce arithmetic cost under standard matrix multiplication kernels.

\noindent\textbf{SVD-based Compression of Large Language Models.}
Singular Value Decomposition (SVD) based compression has recently become a popular approach for LLMs because it is simple, post-training, and can provide inference speedups through explicit low-rank factorization. The most basic approach approximates a weight matrix $W$ by its rank-$r$ truncated SVD, thereby finding a compressed representation which minimizes differences in the reconstructed weights. However, this objective does not account for how the layer is used by the model. FWSVD~\citep{fwsvd} introduced an importance-weighted SVD approach for LLMs, using Fisher information to weight parameters before factorization. Subsequent methods have primarily focused on making SVD compression activation-aware. ASVD~\citep{asvd} rescales weight matrices using diagonal statistics derived from input activation magnitudes prior to SVD truncation. SVD-LLM~\citep{svdllm} instead directly minimizes layerwise activation reconstruction error by incorporating a whitening matrix computed from the Cholesky factorization of the input activation covariance. SVD-LLM further improves accuracy through a small LoRA~\citep{hu2022lora,wang2024lora} residual recovery step. SVD-LLMv2~\citep{svd_llm_V2} extends this framework with heterogeneous layerwise rank allocation based on estimates of truncation-induced loss, and Dobi-SVD~\citep{dobisvd} optimizes layerwise ranks through backpropagation while computing truncated approximations using incremental PCA. Zero-sum SVD \cite{zerosumsvd} directly minimizes activation reconstruction error while pruning singular values such that the predicted loss change is near zero.

Most existing SVD-based LLM compression methods are therefore one-sided: they adapt the low-rank approximation to the input activation distribution, but do not directly model the sensitivity of the network to errors in different output directions. In contrast, a more general second-order view suggests minimizing a double-sided objective captures both the input and output-side or loss-side sensitivity. This form arises naturally when approximating the local loss increase induced by perturbing a weight matrix, since the Hessian or Fisher block associated with a matrix-shaped parameter can often be approximated by Kronecker-factored curvature terms. Prior work has used this idea in double-sided SVD compression utilizing a Kronecker factorization of the network Hessian~\citep{gfwsvd}. Related ideas also appear outside the SVD-compression literature, where Kronecker-factored second-order approximations are used as compression objectives~\citep{yaqa}. These approaches suggest that one-sided activation reconstruction captures only part of the relevant error geometry: two approximations with the same input reconstruction error may have very different effects on the downstream loss depending on the output directions in which the error lies.

Our method builds on this double-sided perspective. Like SVD-LLM~\citep{svdllm,svd_llm_V2}, we perform post-training, activation-aware low-rank compression of LLM weight matrices. However, rather than optimizing only a right-preconditioned reconstruction objective based on input activations, we approximate each layer using a double-sided objective that accounts for both input statistics and output-side sensitivity. This prioritizes directions important to the local model geometry rather than merely explaining weight or activation variance.

\begin{figure*}[!t]
  \centering
  \includegraphics[width=\textwidth, trim=100 100 100 100, clip]{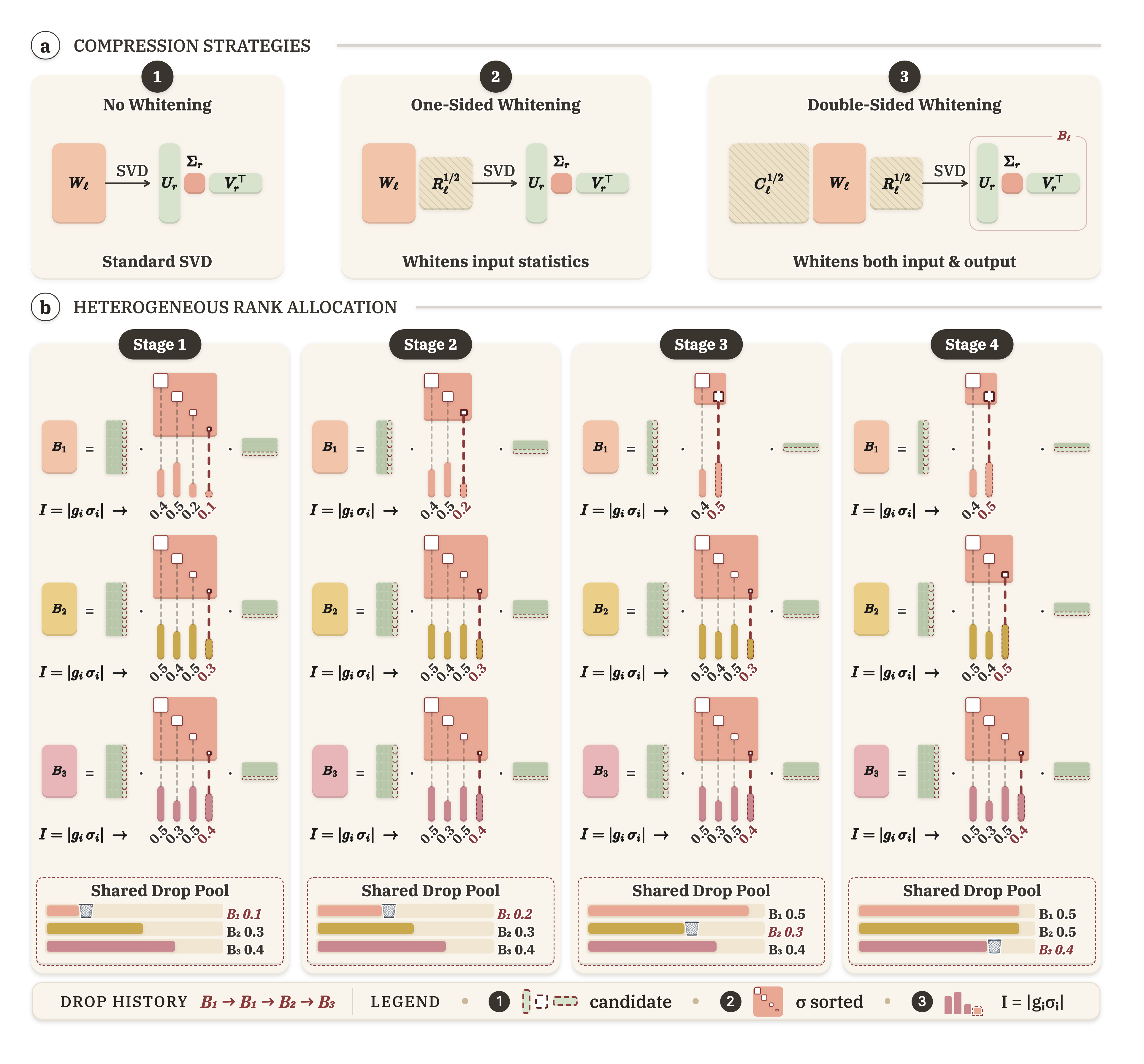}
  \caption{Overview of IO-SVD. \textbf{(a)} Comparison of whitening strategies: standard SVD reconstructs the weight directly, one-sided whitening incorporates only input activation statistics, and double-sided whitening incorporates both input statistics and output-side sensitivity before SVD. \textbf{(b)} Heterogeneous rank allocation. For each whitened matrix $B$, singular components are sorted by singular-value magnitude, and the smallest remaining singular value in each matrix becomes that matrix's drop candidate. These candidates are placed in a shared drop pool and selected by the loss-sensitivity score $I=|g_i\sigma_i|$. In Stage~1, the candidate from $B_1$ has the smallest score and is dropped. In Stage~2, $B_1$ exposes its next-smallest remaining singular value, which is again selected and dropped. In Stage~3, after updating the shared pool, the lowest-score candidate comes from $B_2$ and is removed. Stage~4 shows the resulting drop history, $B_1\!\rightarrow\!B_1\!\rightarrow\!B_2$ ($B_3$ is next), demonstrating that IO-SVD preserves the spectral order within each matrix while producing adaptive ranks across layers.}
  \vspace{-.2in}
  \label{fig:teaser}
\end{figure*}

\vspace{-.1in}
\section{Method}
\label{sec:method}
\vspace{-.1in}

IO-SVD compresses each target linear layer using a KL-aware, input-output whitened SVD objective. The method has three components: (i) a double-sided whitening space derived from the predictive KL divergence between the original and compressed models, (ii) global rank allocation based on singular-component loss sensitivity, and (iii) loss-aware remapping for hybrid SVD--quantization compression.

\vspace{-.1in}
\subsection{KL-aware input-output whitened SVD}
\label{sec:kl_whitened_svd}
\vspace{-.1in}

For a target linear layer \(W_\ell\), let \(h_t=W_\ell x_t\), \(\Delta W_\ell=W_\ell-\hat W_\ell\), and \(\delta h_t=\Delta W_\ell x_t\). Let \(z_t\in\mathbb{R}^{V}\) denote the uncompressed logits at token \(t\), and let \(p_t=\mathrm{softmax}(z_t)\). We seek compressed parameters \(\hat{\theta}\) that preserve the predictive distribution of the original model:
\begin{equation}
\mathcal{J}(\hat{\theta})
=
\mathbb{E}_{(x,y)}\left[
\sum_{t=1}^{T}
\mathrm{KL}\!\left(
p_{\theta}(\cdot \mid x,y_{<t})
\;\|\;
p_{\hat{\theta}}(\cdot \mid x,y_{<t})
\right)\right].
\label{eq:global_kl_objective}
\end{equation}

To obtain a tractable layerwise objective, we approximate the token-level KL around the uncompressed logits with its second order Taylor expansion:
\begin{equation}
\mathrm{KL}\!\left(p_t \,\|\, \mathrm{softmax}(z_t+\delta z_t)\right)
=
\frac{1}{2}\delta z_t^\top H_t\delta z_t
+
O(\|\delta z_t\|^3),
\qquad
H_t=\mathrm{Diag}(p_t)-p_t p_t^\top .
\label{eq:kl_second_order}
\end{equation}
The derivation is given in Appendix~\ref{app:kl_whitening}. Let \(J_t=\partial z_t/\partial h_t\). Using the linearization
\(\delta z_t\approx J_t\Delta W_\ell x_t\), we obtain
\begin{equation}
\Delta\mathcal{J}_{\ell,t}
\approx
\frac{1}{2}
x_t^\top
\Delta W_\ell^\top
C_{\mathrm{token},t}
\Delta W_\ell
x_t,
\qquad
C_{\mathrm{token},t}=J_t^\top H_tJ_t .
\label{eq:token_local_surrogate}
\end{equation}

Averaging over calibration tokens and applying a moment-decoupling approximation yields
\begin{equation}
\Delta\mathcal{J}_{\ell}
\approx
\frac{1}{2}
\left\|
C_\ell^{1/2}
(W_\ell-\hat W_\ell)
R_\ell^{1/2}
\right\|_F^2,
\qquad
R_\ell=\mathbb{E}_t[x_t x_t^\top],
\quad
C_\ell=\mathbb{E}_t[C_{\mathrm{token},t}] .
\label{eq:double_whitened_objective}
\end{equation}
Here, \(R_\ell\) captures input activation geometry, while \(C_\ell\) captures output-side predictive sensitivity. The trace-to-Frobenius reduction is shown in Appendix~\ref{app:trace_to_frobenius}. In practice, we use damped estimates
\(\bar R_\ell=R_\ell+\lambda_R I\) and
\(\bar C_\ell=C_\ell+\lambda_C I\)
when computing matrix square roots and inverse square roots; for notational simplicity, we omit the bars below.

Define the doubly whitened matrix
\begin{equation}
B_\ell=C_\ell^{1/2}W_\ell R_\ell^{1/2}.
\label{eq:whitened_weight}
\end{equation}
With damped positive-definite estimates of \(R_\ell\) and \(C_\ell\), the whitening maps are invertible and preserve rank, so the layerwise compression problem becomes
\begin{equation}
\min_{\operatorname{rank}(\hat B_\ell)\le r_\ell}
\|B_\ell-\hat B_\ell\|_F^2,
\label{eq:whitened_svd_problem}
\end{equation}
where \(\hat B_\ell=C_\ell^{1/2}\hat W_\ell R_\ell^{1/2}\). By the Eckart--Young--Mirsky theorem~\cite{eckart1936approximation,mirsky1960symmetric}, the solution is the rank-\(r_\ell\) truncated SVD
\begin{equation}
\hat B_\ell^\star
=
U_{\ell,r}\Sigma_{\ell,r}V_{\ell,r}^\top .
\label{eq:truncated_svd_solution}
\end{equation}
Undoing the whitening gives
\begin{equation}
\hat W_\ell^\star
=
C_\ell^{-1/2}
U_{\ell,r}\Sigma_{\ell,r}
\left(R_\ell^{-1/2}V_{\ell,r}\right)^\top .
\label{eq:unwhitened_solution}
\end{equation}

\paragraph{Efficiently computing \(C_\ell\).}
Directly forming \(C_{\mathrm{token},t}=J_t^\top H_tJ_t\) is impractical because \(H_t\) and \(J_t\) scale with the vocabulary size. We therefore approximate the KL curvature on the top-\(K\) support of the uncompressed model, renormalize probabilities on this support, and accumulate \(C_\ell\) using vector-Jacobian products with backward hooks. This avoids materializing \(H_t\), \(J_t\), or \(J_t^\top H_tJ_t\). The full top-\(K\) factorization and implementation details are given in Appendix~\ref{app:efficient_curvature}, and an ablation over the choice of $k$ is given in Section~\ref{sec:ablations}. 

\vspace{-.1in}
\subsection{Adaptive rank allocation}
\label{sec:adaptive_rank_allocation}
\vspace{-.1in}

The SVD solution above assumes fixed per-layer ranks. Under a global compression budget, we instead allocate ranks by estimating the calibration-loss impact of removing individual whitened singular components.

Let
$B_\ell
=
U_\ell\Sigma_\ell V_\ell^\top
=
\sum_i
\sigma_{\ell,i}u_{\ell,i}v_{\ell,i}^\top$
, and \(\mathcal{L}\) be the calibration loss, and \(G_\ell=\partial\mathcal{L}/\partial W_\ell\). The corresponding whitened gradient is
\begin{equation}
\widetilde G_\ell
=
C_\ell^{-1/2}G_\ell R_\ell^{-1/2}.
\label{eq:whitened_gradient}
\end{equation}
The first-order derivative with respect to the \(i\)-th singular value is
$g_{\ell,i}=u_{\ell,i}^\top \widetilde G_\ell v_{\ell,i}$.
Dropping this component sets \(\Delta\sigma_{\ell,i}=-\sigma_{\ell,i}\), so we score its importance by
\begin{equation}
I_{\ell,i}=|g_{\ell,i}\sigma_{\ell,i}|.
\label{eq:singular_component_score}
\end{equation}
The derivation of this score is provided in Appendix~\ref{app:rank_score_derivation}.

We greedily remove the lowest-scoring eligible tail component across all layers until the target global parameter budget is reached. After a component is removed from a layer, the next-smallest retained component from that layer becomes eligible, preserving the truncated-SVD structure. The budget is tracked using the layer-specific storage gain defined in Appendix~\ref{sec:iosvd_algorithm}. This procedure induces heterogeneous ranks \(\{r_\ell\}\) without iterative rank optimization.

\vspace{-.1in}
\subsection{Loss-aware remapping for hybrid compression}
\label{sec:loss_aware_remapping}
\vspace{-.1in}

Standard SVD compression ties storage reduction to rank reduction: a rank-\(k\) factorization of \(W\in\mathbb{R}^{m\times n}\) stores \(k(m+n)\) parameters, so aggressive parameter budgets force aggressive spectral truncation. Dobi-SVD~\cite{dobisvd} mitigates this issue by remapping low-rank factors and storing selected rows in low precision. However, its row-selection rule is fixed and structural, rather than based on quantization error or loss impact.

We replace this fixed rule with a loss-aware row-selection criterion. After SVD truncation, write each compressed module as
\begin{equation}
\hat W_\ell=A_\ell D_\ell^\top,
\label{eq:remapping_factorization}
\end{equation}
where \(A_\ell\) and \(D_\ell\) are the retained low-rank factors. Let \(r_{\ell,i}\) denote a candidate row from either factor, and let \(\mathcal{Q}_8(\cdot)\) be the 8-bit quantize-dequantize operator. Quantizing this row induces
\begin{equation}
\Delta r_{\ell,i}
=
\mathcal{Q}_8(r_{\ell,i})-r_{\ell,i}.
\label{eq:row_quantization_perturbation}
\end{equation}
Using calibration data, we compute the row gradient
~$\gamma_{\ell,i}
=
\frac{\partial\mathcal{L}_{\mathrm{cal}}}{\partial r_{\ell,i}}$,
and score the predicted quantization-induced loss by
~$s_{\ell,i}
=
\left|
\left\langle
\gamma_{\ell,i},
\mathcal{Q}_8(r_{\ell,i})-r_{\ell,i}
\right\rangle
\right|$.

Given the remaining compression budget after SVD truncation, we place all candidate rows into a shared pool and greedily select rows with the smallest predicted loss cost, or cost per saved parameter when row sizes differ:
$\sum_{(\ell,i)\in\mathcal{S}} c_{\ell,i}
\ge
C_{\mathrm{rem}}$, and
$C_{\mathrm{rem}}
=
\max\{0, C_{\mathrm{target}}-C_{\mathrm{svd}}\}$.
\label{eq:remapping_budget}

Rows in \(\mathcal{S}\) are stored in 8-bit precision, along with their indices. This preserves the storage benefit of Dobi-SVD-style remapping while making row selection sensitive to both quantization error and task-loss impact. Additional details are given in Appendix~\ref{app:remapping_details}.



\begin{figure*}[t]
  \centering
  \makebox[\textwidth][c]{\includegraphics[width=1.\textwidth]{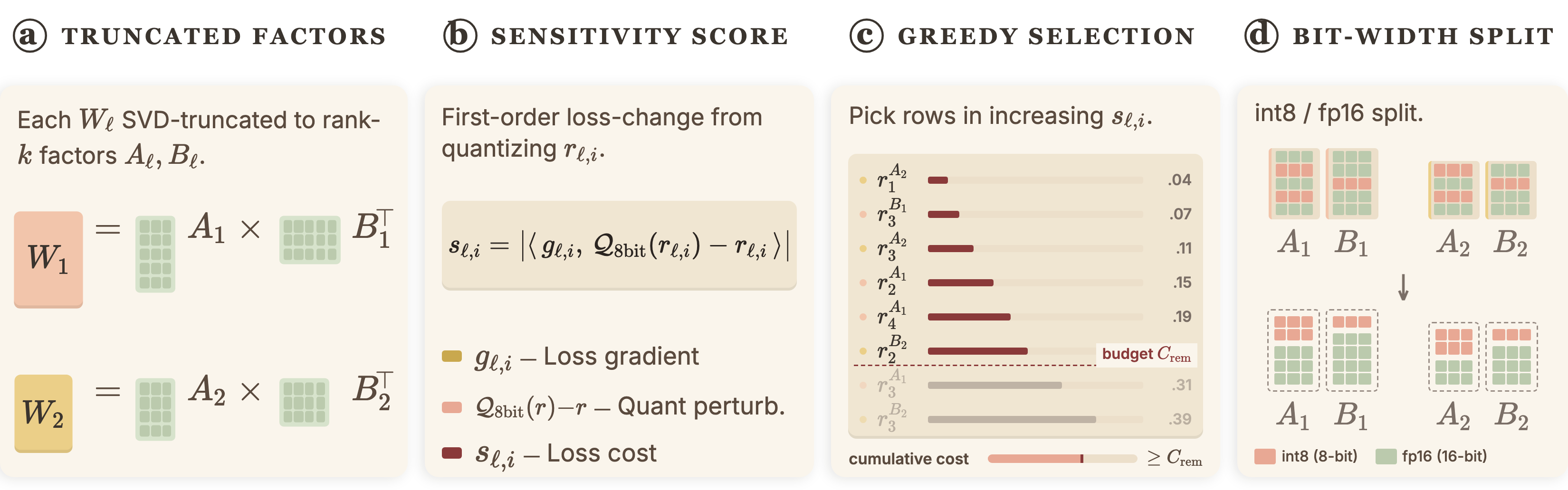}}
  \caption{Loss-aware remapping: \textbf{(a)} SVD-truncate each weight to rank $k$; \textbf{(b)} score factor rows by first-order calibration-loss change under int8 quantization; \textbf{(c)} greedily keep low-score rows in int8 until meeting $C_{\mathrm{rem}}$; \textbf{(d)} assign the remaining rows to fp16.}
  \label{fig:loss_aware_remapping}
  \vspace{-.15in}
\end{figure*}

\begin{table*}[!t]
  \caption{IO-SVD vs.\ SVD-based compression methods on LLaMA-7B across maintenance ratios (0.8/0.6/0.4). Lower is better for PPL; best is in bold. ($^\ast$) indicates results with Dobi-SVD-style remapping enabled. ($^\ddagger$) denotes IO-SVD with our loss-aware remapping. ($^\dagger$) denotes \textsc{HQ} (Half-prune+Quant), which we use in place of remapping at pruning $\ge 50\%$.\vspace{-.15in}}

  \label{tab:zs_svd_main}
  \begin{center}
    \begin{scriptsize}
      \setlength{\tabcolsep}{3pt}
      \renewcommand{\arraystretch}{0.95}
      \resizebox{\textwidth}{!}{%
        \begin{tabular}{c l c c c c c c c c c c c}
          \toprule
          \multirow{2}{*}{Ratio} & \multirow{2}{*}{Method} &
          \multicolumn{3}{c}{PPL $\downarrow$} &
          \multicolumn{7}{c}{Acc $\uparrow$} &
          \multirow{2}{*}{Avg.\ $\uparrow$} \\
          \cmidrule(lr){3-5}\cmidrule(lr){6-12}
          & & Wiki2 & PTB & C4 & Openb. & ARC\_e & ARC\_c & WinoG. & HellaS. & PIQA & MathQA & \\
          \midrule
          1.0 & Baseline
          & 5.68 & 8.35 & 7.34
          & 0.34 & 0.75 & 0.42 & 0.70 & 0.57 & 0.79 & 0.28
          & 0.55 \\
          \midrule
          \multirow{9}{*}{0.8}
          & ASVD
          & 11.14 & 16.55 & 15.93
          & 0.25 & 0.53 & 0.27 & 0.64 & 0.41 & 0.68 & 0.24
          & 0.43 \\
          & SVD-LLM
          & 7.94 & 16.22 & 15.84
          & 0.22 & 0.58 & 0.29 & 0.63 & 0.43 & 0.69 & 0.24
          & 0.44 \\
          & Dobi-SVD
          & 8.54 & 14.83 & 10.01
          & 0.26 & 0.59 & 0.31 & 0.66 & 0.44 & 0.70 & 0.23
          & 0.46 \\
          & ZS-SVD
          & 6.74 & 11.87 & 10.74
          & 0.31 & 0.70 & 0.37 & 0.68 & 0.49 & 0.74 & 0.24
          & 0.50 \\

          & IO-SVD
          & 6.41 & 10.93 & 9.82
          & 0.30 & 0.71 & 0.35 & 0.69 & 0.50 & 0.74 & 0.24
          & 0.50 \\
          \cmidrule(lr){2-13}
          & Dobi-SVD$^\ast$
          & 6.08 & 15.39 & 7.83
          & 0.27 & 0.65 & 0.37 & 0.68 & 0.54 & 0.77 & 0.27
          & 0.51 \\
          & ZS-SVD$^\ast$
          & 5.90 & 8.81 & 7.95
          & 0.35 & 0.74 & 0.41 & 0.70 & 0.56 & 0.78 & 0.26
          & 0.54 \\
          & IO-SVD$^\ddagger$
          & 5.59 & 8.56 & 7.62
          & 0.34 & 0.75 & 0.40 & 0.70 & 0.56 & 0.78 & 0.26
          & 0.54 \\

          \midrule
          \multirow{9}{*}{0.6}
          & ASVD
          & 1407 & 3292 & 1109
          & 0.13 & 0.28 & 0.22 & 0.48 & 0.26 & 0.55 & 0.19
          & 0.30 \\
          & SVD-LLM
          & 13.11 & 63.75 & 49.83
          & 0.19 & 0.42 & 0.25 & 0.58 & 0.33 & 0.60 & 0.21
          & 0.37 \\
          & Dobi-SVD
          & 13.54 & 46.38 & 23.54
          & 0.22 & 0.41 & 0.27 & 0.58 & 0.34 & 0.61 & 0.23
          & 0.38 \\
          & ZS-SVD
          & 11.44 & 43.19 & 34.13
          & 0.23 & 0.52 & 0.26 & 0.62 & 0.35 & 0.64 & 0.22
          & 0.41 \\
          & IO-SVD
          & 9.84 & 28.84 & 27.15
          & 0.22 & 0.55 & 0.25 & 0.62 & 0.37 & 0.64 & 0.22
          & 0.41 \\
          \cmidrule(lr){2-13}
          & Dobi-SVD$^\ast$
          & 8.12 & 43.85 & 12.63
          & 0.28 & 0.65 & 0.32 & 0.62 & 0.45 & 0.72 & 0.25
          & 0.47 \\
          & ZS-SVD$^\ast$
          & 6.96 & 12.72 & 11.52
          & 0.32 & 0.71 & 0.36 & 0.68 & 0.48 & 0.74 & 0.24
          & 0.50 \\
          & IO-SVD$^\ddagger$
          & 6.27 & 10.89 & 10.15
          & 0.31 & 0.69 & 0.35 & 0.68 & 0.50 & 0.74 & 0.26
          & 0.50 \\

          \midrule
          \multirow{9}{*}{0.4}
          & ASVD
          & 57057 & 45218 & 43036
          & 0.12 & 0.26 & 0.21 & 0.49 & 0.26 & 0.53 & 0.18
          & 0.29 \\
          & SVD-LLM
          & 53.74 & 438.58 & 345.49
          & 0.14 & 0.28 & 0.22 & 0.50 & 0.27 & 0.55 & 0.21
          & 0.31 \\
          & Dobi-SVD
          & 46.18 & 238.91 & 190.62
          & 0.15 & 0.31 & 0.20 & 0.52 & 0.28 & 0.54 & 0.22
          & 0.32 \\
          & ZS-SVD
          & 45.17 & 334.85 & 212.57
          & 0.15 & 0.31 & 0.21 & 0.54 & 0.28 & 0.55 & 0.21
          & 0.32 \\

          & IO-SVD
          & 27.70 & 189.53 & 136.42
          & 0.13 & 0.31 & 0.19 & 0.53 & 0.28 & 0.55 & 0.21
          & 0.31 \\
          \cmidrule(lr){2-13}
          & Dobi-SVD$^\ast$
          & 9.95 & 67.62 & 17.94
          & 0.23 & 0.52 & 0.24 & 0.56 & 0.38 & 0.65 & 0.23
          & 0.40 \\
          & ZS-SVD$^\dagger$
          & 6.73 & 11.78 & 10.69
          & 0.29 & 0.72 & 0.38 & 0.68 & 0.49 & 0.75 & 0.25
          & 0.51 \\
          & IO-SVD$^\dagger$
          & 6.41 & 10.82 & 9.79
          & 0.30 & 0.70 & 0.36 & 0.67 & 0.50 & 0.75 & 0.25
          & 0.51 \\

          \bottomrule
        \end{tabular}%
      }
    \end{scriptsize}
  \end{center}
  \vspace{-.25in}
\end{table*}

\vspace{-.1in}
\section{Experiments}
\vspace{-.1in}

\begin{table*}[!t]
\centering
\caption{Accuracy evaluation of different methods under FP16.}
\vspace{-.1in}
\label{tab:vlm_fp16}
\setlength{\tabcolsep}{4pt}
\renewcommand{\arraystretch}{1.15}
\resizebox{\textwidth}{!}{%
\begin{tabular}{cl|ccccc|ccccc|c}
\toprule
\multirow{2}{*}{Model.} & \multirow{2}{*}{Method}
  & \multicolumn{5}{c|}{ScienceQA-IMG $\uparrow$}
  & \multicolumn{5}{c|}{SEED-Bench $\uparrow$}
  & \multirow{2}{*}{Avg.\ $\uparrow$} \\
& & ratio: 90\% & ratio: 80\% & ratio: 70\% & ratio: 60\% & ratio: 50\%
  & ratio: 90\% & ratio: 80\% & ratio: 70\% & ratio: 60\% & ratio: 50\% & \\
\midrule

\multirow{6}{*}{\rotatebox{90}{\shortstack{LLaVA1.5\\7B}}}
 & ASVD       & 49.93 & 50.12 & 47.10 & 36.69 & 19.19 & 54.27 & 53.53 & 48.35 & 37.17 & 24.17 & 42.05 \\
 & SVD-LLM    & 65.44 & 63.71 & 61.92 & 57.41 & 55.53 & 57.89 & 57.50 & 55.33 & 54.64 & 55.31 & 58.47 \\
 & QSVD       & 67.72 & \textbf{68.22} & 67.08 & 65.05 & 62.37 & 59.84 & 59.07 & 59.78 & 59.00 & 58.23 & 62.64 \\
 & WSVD       & \textbf{68.17} & 67.72 & \textbf{67.28} & \textbf{65.89} & \textbf{65.49} & \textbf{60.10} & \textbf{60.17} & \textbf{59.89} & \textbf{60.18} & \textbf{60.46} & \textbf{63.54} \\
 & Ours       & \textbf{67.92} & \textbf{68.07} & \textbf{68.17} & \textbf{66.73} & \textbf{65.99} & \textbf{60.29} & \textbf{60.32} & \textbf{60.22} & \textbf{60.15} & \textbf{60.64} & \textbf{63.85} \\
\cmidrule{2-13}
 & FP16 & \multicolumn{5}{c|}{Accuracy: 68.01} & \multicolumn{5}{c|}{Accuracy: 60.18} & 64.10 \\
\midrule

\multirow{6}{*}{\rotatebox{90}{\shortstack{LLaVA1.5\\13B}}}
 & ASVD       & 71.39 & 71.59 & 70.00 & 70.25 & 69.51 & 61.92 & 61.91 & 61.54 & 61.51 & 60.71 & 66.03 \\
 & SVD-LLM    & 71.05 & 70.85 & 70.30 & 70.35 & 70.30 & 62.28 & 62.34 & 62.25 & 62.08 & \textbf{63.01} & 66.48 \\
 & QSVD       & \textbf{71.89} & \textbf{71.99} & 71.49 & 71.54 & 71.39 & \textbf{62.61} & \textbf{62.64} & \textbf{62.82} & \textbf{62.63} & \textbf{62.52} & 67.15 \\
 & WSVD       & \textbf{71.99} & 71.84 & \textbf{72.53} & \textbf{71.59} & \textbf{71.44} & 62.52 & \textbf{62.68} & 62.38 & 62.37 & 62.37 & \textbf{67.17} \\
 & Ours       & 71.79 & \textbf{72.14} & \textbf{72.29} & \textbf{72.14} & \textbf{72.68} & \textbf{62.64} & 62.49 & \textbf{62.46} & \textbf{62.45} & 62.29 & \textbf{67.34} \\
\cmidrule{2-13}
 & FP16 & \multicolumn{5}{c|}{Accuracy: 71.83} & \multicolumn{5}{c|}{Accuracy: 62.53} & 67.18 \\
\midrule

\multirow{7}{*}{\rotatebox{90}{\shortstack{SmolVLM\\2B}}}
 & & ratio: 90\% & \multicolumn{2}{c}{ratio: 80\%} & \multicolumn{2}{c|}{ratio: 70\%} & ratio: 90\% & \multicolumn{2}{c}{ratio: 80\%} & \multicolumn{2}{c|}{ratio: 70\%} & \\
\cmidrule{3-12}
 & ASVD       & 29.30 & \multicolumn{2}{c}{3.97}  & \multicolumn{2}{c|}{0.20}  & 17.85 & \multicolumn{2}{c}{1.50}  & \multicolumn{2}{c|}{0.95}  & 8.96 \\
 & SVD-LLM    & 40.06 & \multicolumn{2}{c}{17.20} & \multicolumn{2}{c|}{3.82}  & 32.49 & \multicolumn{2}{c}{15.89} & \multicolumn{2}{c|}{4.60}  & 19.01 \\
 & QSVD       & \textbf{77.00} & \multicolumn{2}{c}{62.77} & \multicolumn{2}{c|}{42.59} & 64.80 & \multicolumn{2}{c}{50.46} & \multicolumn{2}{c|}{36.24} & 55.64 \\
 & WSVD       & 76.30 & \multicolumn{2}{c}{\textbf{71.74}} & \multicolumn{2}{c|}{\textbf{60.93}} & \textbf{65.78} & \multicolumn{2}{c}{\textbf{63.29}} & \multicolumn{2}{c|}{\textbf{54.45}} & \textbf{65.42} \\
 & Ours       & \textbf{83.34} & \multicolumn{2}{c}{\textbf{82.65}} & \multicolumn{2}{c|}{\textbf{82.30}} & \textbf{68.36} & \multicolumn{2}{c}{\textbf{68.65}} & \multicolumn{2}{c|}{\textbf{68.15}} & \textbf{75.58} \\
\cmidrule{2-13}
 & FP16 & \multicolumn{5}{c|}{Accuracy: 84.53} & \multicolumn{5}{c|}{Accuracy: 68.47} & 76.53 \\
\bottomrule
\end{tabular}%
}
\vspace{-.25in}
\end{table*}


To evaluate both generative modeling quality and task-level behavior, we report perplexity on WikiText2~\citep{wikitext}, Penn Treebank~\citep{ptb}, and C4~\citep{c4}, as well as zero-shot accuracy on OpenBookQA~\citep{openbookqa}, ARC-Easy, ARC-Challenge~\citep{arc}, WinoGrande~\citep{winogrande}, HellaSwag~\citep{hellaswag}, PIQA~\citep{piqa}, and MathQA~\citep{mathqa}. Following the standard setup in prior SVD-based LLM compression work, we apply compression to the primary transformer linear projections, including the query, key, value, and output attention projections, as well as the MLP layers. For calibration, we use 256 randomly sampled WikiText2 sequences with sequence length 2048, consistent with prior studies~\citep{svdllm, dobisvd, zerosumsvd}.

\vspace{-.1in}
\subsection{Results}
\vspace{-.1in}

Table~\ref{tab:zs_svd_main} compares IO-SVD against SVD-based compression baselines, including ASVD~\citep{asvd}, SVD-LLM~\citep{svdllm}, Dobi-SVD~\citep{dobisvd}, and ZS-SVD~\citep{zerosumsvd}, under maintenance ratios of 80\%, 60\%, and 40\%. Across these settings, IO-SVD consistently improves over the baselines on perplexity, both on in-distribution evaluation and out-of-distribution datasets. On downstream zero-shot commonsense reasoning tasks, IO-SVD also achieves comparable or better accuracy than prior methods.

The results below the horizontal bar in Table~\ref{tab:zs_svd_main} report remapping comparisons, where vanilla remapped variants of existing methods are compared against our loss-aware remapping strategy (denoted by markers in the table). Consistent with the observation in ZS-SVD~\citep{zerosumsvd}, for pruning ratios above 50\% (0.4 ratio in Table~\ref{tab:zs_svd_main} $\rightarrow$ 60\% pruning), we use half-prune quantization (HQ): first applying SVD truncation at twice the target maintenance ratio, then using 8-bit quantization to reach the final budget. A more detailed ablation over remapping strategies is provided in Section~\ref{sec:remapping_ablation}.

Table~\ref{tab:vlm_fp16} reports VLM compression results on LLaVA-1.5 7B, LLaVA-1.5 13B, and SmolVLM 2B. We use 256 randomly sampled ScienceQA-IMG sequences with length 2048 as calibration data, and evaluate the compressed models on ScienceQA-IMG~\citep{lu2022learn_sqa} and SEED-Bench~\citep{li2024seed}. We vary the parameter retention ratio across 90\%, 80\%, 70\%, 60\%, and 50\% where applicable, and compare against ASVD~\citep{asvd}, SVD-LLM~\citep{svdllm}, QSVD~\citep{wangqsvd}, and WSVD~\citep{wangwsvd}. Consistent with QSVD and WSVD, which are VLM-specific compression methods, we apply truncation only to the query, key, and value attention projections. Across models and retention ratios, IO-SVD achieves the best average accuracy among compressed methods, with particularly strong gains on SmolVLM 2B, where it remains close to the FP16 model while substantially outperforming prior SVD-based baselines.

\begin{table*}[t]
\centering
\begin{minipage}[t]{0.52\textwidth}
\centering
\captionof{table}{Perplexity (PPL, $\downarrow$) on WikiText-2 and average accuracy (Acc, $\uparrow$) on six commonsense reasoning datasets (excluding arc\_c) for OPT-6.7B, Vicuna-7B, and LLaMA-13B at 20\% pruning.\vspace{-.05in}}
\label{tab:cross_arch_30b}
\scriptsize
\setlength{\tabcolsep}{3pt}
\renewcommand{\arraystretch}{0.95}
\begin{tabular}{l c c c c c c}
\toprule
& \multicolumn{2}{c}{\textbf{OPT-6.7B}} & \multicolumn{2}{c}{\textbf{Vicuna-7B}} & \multicolumn{2}{c}{\textbf{LLaMA-13B}} \\
\cmidrule(lr){2-3} \cmidrule(lr){4-5} \cmidrule(lr){6-7}
\textbf{Method} & \textbf{PPL} & \textbf{Acc} & \textbf{PPL} & \textbf{Acc} & \textbf{PPL} & \textbf{Acc} \\
\midrule
Original  & 10.86 & 0.52 & 6.78 & 0.56 & 5.09 & 0.59 \\
\cmidrule(lr){1-7}
SVD       & 66275 & 0.03 & 18644 & 0.05 & 946.31 & 0.21 \\
FWSVD     & 14559 & 0.06 & 2758  & 0.09 & 15.98 & 0.43 \\
ASVD      & 82.00 & 0.32 & 16.23 & 0.33 & 6.74 & 0.54 \\
SVDLLM    & 16.04 & 0.41 & 8.41 & 0.51 & 6.43 & 0.55 \\
ZS-SVD    & 11.40 & 0.51 & 8.08 & \textbf{0.54} & 5.84 & 0.56 \\
IO-SVD    & \textbf{11.10} & \textbf{0.51} & \textbf{7.36} & 0.53 & \textbf{5.60} & \textbf{0.56} \\
\bottomrule
\end{tabular}


\captionof{table}{Ablation over whitening curvature and rank allocation on LLaMA-7B, calibrated on WikiText2. WikiText2 PPL across pruning ratios; lower is better.\vspace{.0in}}
\label{tab:whitening_rank_ablation}
\scriptsize
\setlength{\tabcolsep}{1.5pt}
\renewcommand{\arraystretch}{0.95}
\begin{tabular}{l c c c c}
\toprule
\textbf{Curvature} & \textbf{Het. rank?} & \textbf{Ratio 0.8} & \textbf{Ratio 0.6} & \textbf{Ratio 0.4} \\
\midrule
Input-only (SVD-LLM)   & \ding{55} & 7.95 & 13.11 & 53.74 \\
Double-sided (OBD-LLM) & \ding{55} & 7.36 & 11.34 & 32.95 \\
Double-sided (Ours)    & \ding{55} & 7.31 & 11.20 & 32.09 \\
\cmidrule(lr){1-5}
Input-only (SVD-LLM)   & \ding{51} & 6.72 & 11.65 & 62.76 \\
Double-sided (OBD-LLM) & \ding{51} & 6.45 & 9.90 & 28.19 \\
Double-sided (Ours)    & \ding{51} & \textbf{6.41} & \textbf{9.84} & \textbf{27.70} \\
\bottomrule
\end{tabular}
\end{minipage}%
\hfill
\begin{minipage}[t]{0.46\textwidth}
\centering
\captionof{table}{Commonsense reasoning results on LLaMA-2-7B. IO-SVD and its remapped variants are compared with pruning and SVD-based baselines at compression ratios 0.6 and 0.4, higher is better.} \vspace{.015in}
\label{tab:llama2_7b_commonsense}
\scriptsize
\setlength{\tabcolsep}{2pt}
\renewcommand{\arraystretch}{0.95}
\begin{tabular}{c l c c c c c c}
\toprule
\multirow{2}{*}{\textbf{Ratio}} & \multirow{2}{*}{\textbf{Method}} &
\multicolumn{5}{c}{\textbf{Acc $\uparrow$}} &
\multirow{2}{*}{\textbf{Avg\ $\uparrow$}}  \\
\cmidrule(lr){3-7}
& & \textbf{PIQA} & \textbf{Hell} & \textbf{Win} & \textbf{ARC\_e} & \textbf{ARC\_c}  \\
\midrule
1.0 & Baseline & 0.78 & 0.57 & 0.69 & 0.76 & 0.43 & 0.65 \\
\midrule
\multirow{10}{*}{0.6} & LLM-Pruner & 0.70 & 0.41 & 0.53 & 0.53 & 0.27 & 0.48 \\
& SliceGPT & 0.65 & 0.57 & 0.60 & 0.43 & 0.32 & 0.51  \\
& Bonsai & 0.72 & 0.45 & 0.58 & 0.59 & 0.30 & \textbf{0.53}   \\
& Wanda-sp & 0.70 & 0.42 & 0.53 & 0.57 & 0.29 & 0.50  \\
& SVD-LLM & 0.56 & 0.30 & 0.57 & 0.39 & 0.21 & 0.41  \\
& ZS-SVD & 0.63 & 0.34 & 0.60 & 0.46 & 0.25 & 0.45  \\
& IO-SVD & 0.61 & 0.33 & 0.59 & 0.51 & 0.23 & 0.45 \\
\cmidrule(lr){2-8}
& Dobi-SVD$^\ast$ & 0.72 & 0.45 & 0.64 & 0.67 & 0.31 & 0.56 \\
& ZS-SVD$^\ast$ & 0.72 & 0.46 & 0.67 & 0.66 & 0.33 & 0.57 \\
& IO-SVD$^\ddagger$ & 0.74 & 0.47 & 0.67 & 0.73 & 0.38 & \textbf{0.60} \\
\midrule
\multirow{6}{*}{0.4} & SVD-LLM & 0.54 & 0.27 & 0.48 & 0.26 & 0.20 & 0.35  \\
& ZS-SVD & 0.54 & 0.27 & 0.52 & 0.29 & 0.19 & 0.36 \\
& IO-SVD & 0.54 & 0.28 & 0.52 & 0.30 & 0.19 & \textbf{0.37} \\
\cmidrule(lr){2-8}
& Dobi-SVD$^\ast$ & 0.67 & 0.38 & 0.57 & 0.55 & 0.26 & 0.49 \\
& ZS-SVD$^\dagger$ & 0.73 & 0.48 & 0.68 & 0.70 & 0.36 & \textbf{0.59} \\
& IO-SVD$^\dagger$ & 0.73 & 0.48 & 0.65 & 0.71 & 0.37 & \textbf{0.59} \\
\bottomrule
\end{tabular}
\end{minipage}
\vskip -0.1in
\end{table*}

Tables~\ref{tab:cross_arch_30b} and~\ref{tab:llama2_7b_commonsense} further evaluate IO-SVD across model families and pruning settings. In the cross-architecture study (Table~\ref{tab:cross_arch_30b}), we test OPT-6.7B, Vicuna-7B, and LLaMA-13B at 20\% pruning, covering different LLM families and model scales. IO-SVD achieves the lowest perplexity across all three models and matches or improves the average commonsense accuracy compared with the strongest SVD-based baselines, showing that the method is not tied to a single architecture. Table~\ref{tab:llama2_7b_commonsense} evaluates LLaMA-2-7B on commonsense reasoning benchmarks and compares against both pruning and SVD-based compression methods. At the reported compression ratios, vanilla IO-SVD performs better than or on par with prior SVD methods, while the loss-aware remapping variant substantially outperforms pruning baselines in average accuracy.


\vspace{-.1in}
\subsection{Ablation studies}
\label{sec:ablations}
\vspace{-.1in}

\begin{figure}[!t]
  \centering
  \begin{minipage}[c]{0.28\textwidth}
    \centering
    \includegraphics[width=\linewidth]{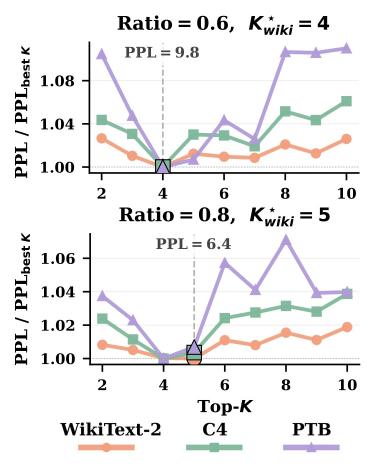}
    \vspace{-.3in}
    \captionof{figure}{Top-K ablation for output-side KL curvature. Normalized perplexity on wiki2, C4, PTB}
    \label{fig:jacobian_topk}
  \end{minipage}%
  \hspace{0.01\textwidth}%
  \begin{minipage}[c]{0.70\textwidth}
    \centering
    \captionof{table}{Effect of standard and loss-aware remapping on LLaMA-7B. Perplexity on wiki2, C4, and PTB at compression ratios 0.8 and 0.6, lower is better.}
    \label{tab:remap_ablation}
    \footnotesize
    \setlength{\tabcolsep}{2.5pt}
    \renewcommand{\arraystretch}{0.98}
    \begin{tabular}{ll ccc ccc}
    \toprule
    \multirow{2}{*}{\textbf{Method}}
      & \multirow{2}{*}{\textbf{Quant mode}}
      & \multicolumn{3}{c}{\textbf{Ratio 0.8}}
      & \multicolumn{3}{c}{\textbf{Ratio 0.6}} \\
    \cmidrule(lr){3-5}\cmidrule(lr){6-8}
    & & \textbf{Wiki}$\downarrow$ & \textbf{C4}$\downarrow$ & \textbf{PTB}$\downarrow$
      & \textbf{Wiki}$\downarrow$ & \textbf{C4}$\downarrow$ & \textbf{PTB}$\downarrow$ \\
    \midrule
    \multirow{3}{*}{SVD-LLM}
      & compressed                       & 7.94 & 15.84 & 16.22 & 13.11 & 49.83 & 63.75 \\
      & \ $+$ remap$^\ast$                & 5.86 &  7.82 &  8.82 &  6.98 & 11.59 &  12.88 \\
      & \ $+$ loss-aware$^\ddagger$       & \textbf{5.66} & \textbf{7.78} & \textbf{8.71} & \textbf{6.69} & \textbf{11.39} & \textbf{12.46} \\
    \cmidrule(lr){1-8}
    \multirow{3}{*}{ZS-SVD}
      & compressed                       & 6.74 & 10.74 & 11.87 & 11.44 & 34.13 & 43.19 \\
      & \ $+$ remap$^\ast$                & 5.90 &  7.95 &  8.81 &  6.96 & 11.52 & \textbf{12.72} \\
      & \ $+$ loss-aware$^\ddagger$       & \textbf{5.69} & \textbf{7.92} & \textbf{8.78} & \textbf{6.69} & \textbf{11.46} & 12.80 \\
    \cmidrule(lr){1-8}
    \multirow{3}{*}{IO-SVD (Ours)}
      & compressed                       & 6.41 &  9.82 & 10.93 &  9.84 & 27.15 & 28.84 \\
      & \ $+$ remap$^\ast$                & 5.76 & \textbf{7.61} &  8.59 &  6.48 & 10.24 & 10.95 \\
      & \ $+$ loss-aware$^\ddagger$       & \textbf{5.59} & 7.62 & \textbf{8.56} & \textbf{6.27} & \textbf{10.15} & \textbf{10.89} \\
    \bottomrule
    \end{tabular}
  \end{minipage}
  \vspace{-.25in}
\end{figure}

{\bf Effect of top-$k$:}
Figure~\ref{fig:jacobian_topk} ablates the effect of the top-$K$ value used to construct the output-side KL curvature. Specifically, we sweep $K$ using WikiText-2 as the selection set, then evaluate the resulting compressed models on the evaluation sets of WikiText-2, PTB, and C4. For better visibility, the plot reports normalized perplexity on the y-axis, measured as the ratio between the perplexity at each $K$ and the best perplexity obtained on WikiText-2. We observe a clear sweet spot for $K$. Importantly, the best $K$ selected on WikiText-2 transfers well across datasets, suggesting that the top-K curvature estimate generalizes across evaluation datasets.


{\bf Whitening space and rank allocation:}
\label{sec:whitening_rank_ablation}
Table~\ref{tab:whitening_rank_ablation} studies the effect of the whitening space and rank-allocation strategy. We compare three whitening choices: input-only whitening as in SVD-LLM, double-sided Kronecker whitening as in OBD-LLM, and our double-sided KL whitening, each under homogeneous and heterogeneous rank allocation. Heterogeneous allocation improves performance in most settings, indicating that a fixed rank assignment is generally suboptimal when different layers exhibit different sensitivity to compression. Among the whitening spaces, double-sided methods consistently outperform input-only whitening, especially at more aggressive pruning ratios, highlighting the benefit of incorporating output-side sensitivity. The best results are achieved by combining heterogeneous rank allocation with our double-sided KL whitening, which obtains the lowest perplexity across all reported compression settings.



{\bf Remapping applied to our method:}
\label{sec:remapping_ablation}
Table~\ref{tab:remap_ablation} evaluates remapping on LLaMA-7B at compression ratios of 80\% and 60\% (pruning rates: 20\% and 40\%). We evaluate on WikiText2, C4, and PTB, and compare vanilla compression, standard remapping, and our loss-aware remapping across SVD-LLM, ZS-SVD, and IO-SVD. Consistent with prior findings on hybrid SVD-quantization compression~\citep{dobisvd, zerosumsvd}, remapping improves over the vanilla compressed models, showing that selective quantization helps recover performance after SVD truncation. Our loss-aware remapping further improves over standard remapping in most settings by choosing rows based on predicted calibration-loss impact, with IO-SVD achieving the best or near-best perplexity across datasets and compression ratios.

\subsubsection{Inference speed and memory}
\label{sec:inference_speed}

\begin{wrapfigure}{r}{0.45\textwidth}
\centering
\includegraphics[width=\linewidth]{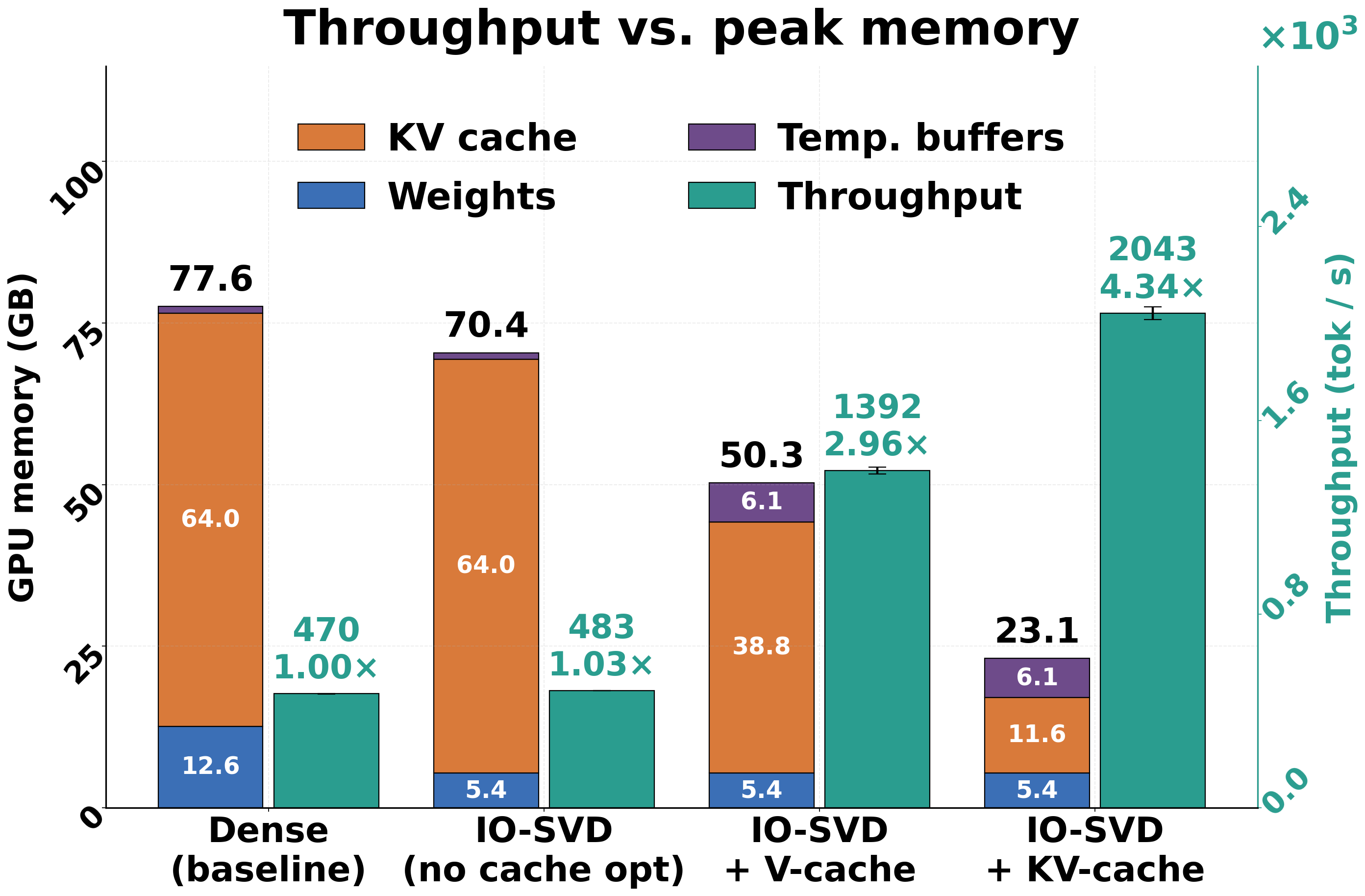}
\vspace{-1.4em}
\caption{Throughput vs.\ peak memory on LLaMA-2-7B (batch 64, seq 1024+1024).}
\label{fig:inference_speed}
\vspace{-1.0em}
\end{wrapfigure}

Figure~\ref{fig:inference_speed} reports decode-time efficiency of IO-SVD against the dense baseline at the same compression budget used in our perplexity and accuracy experiments. We measure both end-to-end throughput in tokens per second and the peak GPU memory footprint, broken down into weight storage, KV cache, and other runtime activations. All inference experiments are conducted on a single NVIDIA RTX PRO 6000 Blackwell Max-Q workstation GPU with 96~GB VRAM.

\textbf{Decode throughput.} IO-SVD without any KV-cache optimization already removes a portion of the per-step compute and yields a small improvement over the dense baseline (483 vs.\ 470~tok/s, $1.03\times$). Combining IO-SVD with V-cache compression raises throughput to 1392~tok/s ($2.96\times$), and adding KV-cache compression pushes it further to 2043~tok/s ($4.34\times$). The progression demonstrates that the speedup of IO-SVD is amplified once memory bandwidth on the cache is reduced, rather than coming purely from the lower-rank weight matrices.

\vspace{1in}

\textbf{Peak GPU memory.} The dense baseline consumes 77.6~GB, dominated by a 64.0~GB KV cache and a 12.6~GB weight tensor. IO-SVD without cache optimization shrinks the weight footprint to 5.4~GB but leaves the KV cache untouched at 64.0~GB, giving a 70.4~GB peak. Adding V-cache compression reduces the cache to 38.8~GB and yields a 50.3~GB peak, while combining V- and KV-cache compression reduces the cache further to 11.6~GB, for a 23.1~GB total. Across all four configurations, the weight memory remains 5.4~GB, indicating that the additional savings come almost entirely from the cache. Together, these results show that IO-SVD is not only accuracy-competitive but also delivers practical inference speedups and substantially smaller memory footprints, making it amenable to deployment on resource-constrained hardware.


\vspace{.2in}
{\bf Memory compression for decoding speed:}
IO-SVD can reduce inference-time memory traffic by compressing the KV cache. For a compressed
key or value projection in layer $\ell$, let $\hat{W}_{\ell}=A_{\ell}D_{\ell}^{\top}$, following
the notation in Section~3.3. Instead of storing the full-dimensional projection
$\hat{h}_t=\hat{W}_{\ell}x_t=A_{\ell}(D_{\ell}^{\top}x_t)$, we cache the low-dimensional latent
vector $z_t=D_{\ell}^{\top}x_t$, whose dimension is the retained rank $r_{\ell}$.
This reduces KV-cache size and memory-bandwidth requirements during autoregressive decoding,
where each new token requires reading the accumulated cache. Although this introduces additional
compute for the up-projection through $A_{\ell}$ during attention, the tradeoff is favorable when
decoding is memory-bandwidth bound, enabling practical speedups in long-context or
memory-constrained settings. Figure~\ref{fig:inference_speed} shows the throughput and memory footprint of the model when compressing V or KV cache.

\FloatBarrier

\section{Conclusion and limitations}
We presented IO-SVD, a post-training compression method that uses KL-aware double-sided whitening, adaptive rank allocation, and loss-aware remapping to improve SVD-based LLM and VLM compression. Experiments show that IO-SVD better preserves perplexity and downstream accuracy than prior SVD-based methods, especially under aggressive compression. Future work should study calibration sensitivity, improve curvature-estimation efficiency, and evaluate IO-SVD across broader models, tasks, and deployment settings.

IO-SVD has several limitations that suggest useful future directions. First, the method estimates output-side curvature using a top-K token approximation, which may not fully capture sensitivity from the long tail of the vocabulary. Second, the adaptive rank allocation is based on a greedy component-removal strategy; while efficient, it may not always find the globally optimal rank distribution. Finally, IO-SVD is evaluated up to 13B-parameter models, so further validation on larger-scale models would strengthen the evidence for scalability.

\clearpage

\bibliographystyle{plainnat}
\bibliography{references}

@article{eckart1936approximation,
  title={The approximation of one matrix by another of lower rank},
  author={Eckart, Carl and Young, Gale},
  journal={Psychometrika},
  volume={1},
  number={3},
  pages={211--218},
  year={1936},
  publisher={Springer}
}

@article{mirsky1960symmetric,
  title={Symmetric gauge functions and unitarily invariant norms},
  author={Mirsky, Leon},
  journal={The Quarterly Journal of Mathematics},
  volume={11},
  number={1},
  pages={50--59},
  year={1960},
  publisher={Oxford University Press}
}

@inproceedings{OBC_2022,
author = {Frantar, Elias and Singh, Sidak Pal and Alistarh, Dan},
title = {Optimal brain compression: a framework for accurate post-training quantization and pruning},
year = {2022},
isbn = {9781713871088},
publisher = {Curran Associates Inc.},
address = {Red Hook, NY, USA},
booktitle = {Proceedings of the 36th International Conference on Neural Information Processing Systems},
articleno = {323},
numpages = {14},
location = {New Orleans, LA, USA},
series = {NIPS '22}
}

@inproceedings{sparsegpt,
author = {Frantar, Elias and Alistarh, Dan},
title = {SparseGPT: massive language models can be accurately pruned in one-shot},
year = {2023},
publisher = {JMLR.org},
booktitle = {Proceedings of the 40th International Conference on Machine Learning},
articleno = {414},
numpages = {15},
location = {Honolulu, Hawaii, USA},
series = {ICML'23}
}

@article{llmpruner,
  title={{LLM-Pruner}: On the Structural Pruning of Large Language Models},
  author={Ma, Xinyin and Fang, Gongfan and Wang, Xinchao},
  journal={Advances in Neural Information Processing Systems},
  volume={36},
  pages={21702--21720},
  year={2023}
}

@inproceedings{
sun2024a,
title={A Simple and Effective Pruning Approach for Large Language Models},
author={Mingjie Sun and Zhuang Liu and Anna Bair and J Zico Kolter},
booktitle={The Twelfth International Conference on Learning Representations},
year={2024},
url={https://openreview.net/forum?id=PxoFut3dWW}
}

@inproceedings{llm_int8,
author = {Dettmers, Tim and Lewis, Mike and Belkada, Younes and Zettlemoyer, Luke},
title = {LLM.int8(): 8-bit matrix multiplication for transformers at scale},
year = {2022},
isbn = {9781713871088},
publisher = {Curran Associates Inc.},
address = {Red Hook, NY, USA},
booktitle = {Proceedings of the 36th International Conference on Neural Information Processing Systems},
articleno = {2198},
numpages = {15},
location = {New Orleans, LA, USA},
series = {NIPS '22}
}

@inproceedings{
quip,
title={Qu{IP}: 2-Bit Quantization of Large Language Models With Guarantees},
author={Jerry Chee and Yaohui Cai and Volodymyr Kuleshov and Christopher De Sa},
booktitle={Thirty-seventh Conference on Neural Information Processing Systems},
year={2023},
url={https://openreview.net/forum?id=xrk9g5vcXR}
}

@inproceedings{quip_2,
author = {Tseng, Albert and Chee, Jerry and Sun, Qingyao and Kuleshov, Volodymyr and De Sa, Christopher},
title = {QuIP\#: even better LLM quantization with hadamard incoherence and lattice codebooks},
year = {2024},
publisher = {JMLR.org},
booktitle = {Proceedings of the 41st International Conference on Machine Learning},
articleno = {1987},
numpages = {27},
location = {Vienna, Austria},
series = {ICML'24}
}

@inproceedings{
gptq,
title={{OPTQ}: Accurate Quantization for Generative Pre-trained Transformers},
author={Elias Frantar and Saleh Ashkboos and Torsten Hoefler and Dan Alistarh},
booktitle={The Eleventh International Conference on Learning Representations },
year={2023},
url={https://openreview.net/forum?id=tcbBPnfwxS}
}

@inproceedings{smooth_quant,
author = {Xiao, Guangxuan and Lin, Ji and Seznec, Mickael and Wu, Hao and Demouth, Julien and Han, Song},
title = {SmoothQuant: accurate and efficient post-training quantization for large language models},
year = {2023},
publisher = {JMLR.org},
booktitle = {Proceedings of the 40th International Conference on Machine Learning},
articleno = {1585},
numpages = {13},
location = {Honolulu, Hawaii, USA},
series = {ICML'23}
}

@inproceedings{
fwsvd,
title={Language model compression with weighted low-rank factorization},
author={Yen-Chang Hsu and Ting Hua and Sungen Chang and Qian Lou and Yilin Shen and Hongxia Jin},
booktitle={International Conference on Learning Representations},
year={2022},
url={https://openreview.net/forum?id=uPv9Y3gmAI5}
}

@inproceedings{svd_llm_V2,
    title = "{SVD}-{LLM} V2: Optimizing Singular Value Truncation for Large Language Model Compression",
    author = "Wang, Xin  and
      Alam, Samiul  and
      Wan, Zhongwei  and
      Shen, Hui  and
      Zhang, Mi",
    editor = "Chiruzzo, Luis  and
      Ritter, Alan  and
      Wang, Lu",
    booktitle = "Proceedings of the 2025 Conference of the Nations of the Americas Chapter of the Association for Computational Linguistics: Human Language Technologies (Volume 1: Long Papers)",
    month = apr,
    year = "2025",
    address = "Albuquerque, New Mexico",
    publisher = "Association for Computational Linguistics",
    url = "https://aclanthology.org/2025.naacl-long.217/",
    doi = "10.18653/v1/2025.naacl-long.217",
    pages = "4287--4296",
    ISBN = "979-8-89176-189-6"
}

@misc{asvd,
      title={ASVD: Activation-aware Singular Value Decomposition for Compressing Large Language Models}, 
      author={Zhihang Yuan and Yuzhang Shang and Yue Song and Dawei Yang and Qiang Wu and Yan Yan and Guangyu Sun},
      year={2025},
      eprint={2312.05821},
      archivePrefix={arXiv},
      primaryClass={cs.CL},
      url={https://arxiv.org/abs/2312.05821}, 
}

@inproceedings{
svdllm,
title={{SVD}-{LLM}: Truncation-aware Singular Value Decomposition for Large Language Model Compression},
author={Xin Wang and Yu Zheng and Zhongwei Wan and Mi Zhang},
booktitle={The Thirteenth International Conference on Learning Representations},
year={2025},
url={https://openreview.net/forum?id=LNYIUouhdt}
}

@inproceedings{
hu2022lora,
title={Lo{RA}: Low-Rank Adaptation of Large Language Models},
author={Edward J Hu and yelong shen and Phillip Wallis and Zeyuan Allen-Zhu and Yuanzhi Li and Shean Wang and Lu Wang and Weizhu Chen},
booktitle={International Conference on Learning Representations},
year={2022},
url={https://openreview.net/forum?id=nZeVKeeFYf9}
}

@article{wang2024lora,
  title={Lora-ga: Low-rank adaptation with gradient approximation},
  author={Wang, Shaowen and Yu, Linxi and Li, Jian},
  journal={Advances in Neural Information Processing Systems},
  volume={37},
  pages={54905--54931},
  year={2024}
}

@inproceedings{
dobisvd,
title={Dobi-{SVD}: Differentiable {SVD} for {LLM} Compression and Some New Perspectives},
author={Wang Qinsi and Jinghan Ke and Masayoshi Tomizuka and Kurt Keutzer and Chenfeng Xu},
booktitle={The Thirteenth International Conference on Learning Representations},
year={2025},
url={https://openreview.net/forum?id=kws76i5XB8}
}

@misc{gfwsvd,
      title={Generalized Fisher-Weighted SVD: Scalable Kronecker-Factored Fisher Approximation for Compressing Large Language Models}, 
      author={Viktoriia Chekalina and Daniil Moskovskiy and Daria Cherniuk and Maxim Kurkin and Andrey Kuznetsov and Evgeny Frolov},
      year={2025},
      eprint={2505.17974},
      archivePrefix={arXiv},
      primaryClass={cs.LG},
      url={https://arxiv.org/abs/2505.17974}, 
}

@misc{yaqa,
      title={Model-Preserving Adaptive Rounding}, 
      author={Albert Tseng and Zhaofeng Sun and Christopher De Sa},
      year={2025},
      eprint={2505.22988},
      archivePrefix={arXiv},
      primaryClass={cs.LG},
      url={https://arxiv.org/abs/2505.22988}, 
}

@misc{zerosumsvd,
      title={Zero Sum SVD: Balancing Loss Sensitivity for Low Rank LLM Compression},
      author={Ali Abbasi and Chayne Thrash and Haoran Qin and Shansita Sharma and Sepehr Seifi and Soheil Kolouri},
      year={2026},
      eprint={2602.02848},
      archivePrefix={arXiv},
      primaryClass={cs.LG},
      url={https://arxiv.org/abs/2602.02848},
}

@inproceedings{shaoomniquant,
      title={OmniQuant: Omnidirectionally Calibrated Quantization for Large Language Models},
      author={Shao, Wenqi and Chen, Mengzhao and Zhang, Zhaoyang and Xu, Peng and Zhao, Lirui and Li, Zhiqian and Zhang, Kaipeng and Gao, Peng and Qiao, Yu and Luo, Ping},
      booktitle={The Twelfth International Conference on Learning Representations},
}

@article{park2022lut,
      title={LUT-GEMM: Quantized matrix multiplication based on LUTs for efficient inference in large-scale generative language models},
      author={Park, Gunho and Park, Baeseong and Kim, Minsub and Lee, Sungjae and Kim, Jeonghoon and Kwon, Beomseok and Kwon, Se Jung and Kim, Byeongwook and Lee, Youngjoo and Lee, Dongsoo},
      journal={arXiv preprint arXiv:2206.09557},
      year={2022},
}

@article{zhao2024atom,
      title={Atom: Low-bit quantization for efficient and accurate LLM serving},
      author={Zhao, Yilong and Lin, Chien-Yu and Zhu, Kan and Ye, Zihao and Chen, Lequn and Zheng, Size and Ceze, Luis and Krishnamurthy, Arvind and Chen, Tianqi and Kasikci, Baris},
      journal={Proceedings of Machine Learning and Systems},
      volume={6},
      pages={196--209},
      year={2024},
}

@article{lin2405qserve,
      title={QServe: W4A8KV4 quantization and system co-design for efficient LLM serving},
      author={Lin, Yujun and Tang, Haotian and Yang, Shang and Zhang, Zhekai and Xiao, Guangxuan and Gan, Chuang and Han, Song},
      journal={arXiv preprint arXiv:2405.04532},
      year={2024},
}

@inproceedings{frantar2025marlin,
      title={Marlin: Mixed-precision auto-regressive parallel inference on large language models},
      author={Frantar, Elias and Castro, Roberto L and Chen, Jiale and Hoefler, Torsten and Alistarh, Dan},
      booktitle={Proceedings of the 30th ACM SIGPLAN Annual Symposium on Principles and Practice of Parallel Programming},
      pages={239--251},
      year={2025},
}

@article{cheng2024survey,
      title={A survey on deep neural network pruning: Taxonomy, comparison, analysis, and recommendations},
      author={Cheng, Hongrong and Zhang, Miao and Shi, Javen Qinfeng},
      journal={IEEE Transactions on Pattern Analysis and Machine Intelligence},
      volume={46},
      number={12},
      pages={10558--10578},
      year={2024},
      publisher={IEEE},
}

@article{hinton2015distilling,
      title={Distilling the knowledge in a neural network},
      author={Hinton, Geoffrey and Vinyals, Oriol and Dean, Jeff},
      journal={arXiv preprint arXiv:1503.02531},
      year={2015},
}

@inproceedings{gu2024minillm,
      title={MiniLLM: Knowledge distillation of large language models},
      author={Gu, Yuxian and Dong, Li and Wei, Furu and Huang, Minlie},
      booktitle={The Twelfth International Conference on Learning Representations},
      year={2024},
}

@inproceedings{agarwal2024policy,
      title={On-policy distillation of language models: Learning from self-generated mistakes},
      author={Agarwal, Rishabh and Vieillard, Nino and Zhou, Yongchao and Stanczyk, Piotr and Garea, Sabela Ramos and Geist, Matthieu and Bachem, Olivier},
      booktitle={The Twelfth International Conference on Learning Representations},
      year={2024},
}

@article{rao2023parameter,
      title={Parameter-efficient and student-friendly knowledge distillation},
      author={Rao, Jun and Meng, Xv and Ding, Liang and Qi, Shuhan and Liu, Xuebo and Zhang, Min and Tao, Dacheng},
      journal={IEEE Transactions on Multimedia},
      volume={26},
      pages={4230--4241},
      year={2023},
      publisher={IEEE},
}

@article{nguyen2026ctpd,
      title={CTPD: Cross Tokenizer Preference Distillation},
      author={Nguyen, Truong and Van Dat, Phi and Nguyen, Ngan and Van, Linh Ngo and Le, Trung and Nguyen, Thanh Hong},
      journal={arXiv preprint arXiv:2601.11865},
      year={2026},
}

@inproceedings{shinovercoming,
      title={Overcoming Vocabulary Mismatch: Vocabulary-agnostic Teacher Guided Language Modeling},
      author={Shin, Haebin and Ji, Lei and Liu, Xiao and Gong, Yeyun},
      booktitle={Forty-second International Conference on Machine Learning},
}

@article{li2026optimal,
      title={Optimal Brain Decomposition for Accurate LLM Low-Rank Approximation},
      author={Li, Yuhang and Lee, Donghyun and Yin, Ruokai and Panda, Priyadarshini},
      journal={arXiv preprint arXiv:2604.00821},
      year={2026},
}

@inproceedings{
wikitext,
title={Pointer Sentinel Mixture Models},
author={Stephen Merity and Caiming Xiong and James Bradbury and Richard Socher},
booktitle={International Conference on Learning Representations},
year={2017},
url={https://openreview.net/forum?id=Byj72udxe}
}

@article{ptb,
    title = "Building a Large Annotated Corpus of {E}nglish: The {P}enn {T}reebank",
    author = "Marcus, Mitchell P.  and
      Santorini, Beatrice  and
      Marcinkiewicz, Mary Ann",
    editor = "Hirschberg, Julia",
    journal = "Computational Linguistics",
    volume = "19",
    number = "2",
    year = "1993",
    address = "Cambridge, MA",
    publisher = "MIT Press",
    url = "https://aclanthology.org/J93-2004/",
    pages = "313--330"
}

@article{c4,
author = {Raffel, Colin and Shazeer, Noam and Roberts, Adam and Lee, Katherine and Narang, Sharan and Matena, Michael and Zhou, Yanqi and Li, Wei and Liu, Peter J.},
title = {Exploring the limits of transfer learning with a unified text-to-text transformer},
year = {2020},
issue_date = {January 2020},
publisher = {JMLR.org},
volume = {21},
number = {1},
issn = {1532-4435},
journal = {J. Mach. Learn. Res.},
month = jan,
articleno = {140},
numpages = {67}
}

@inproceedings{openbookqa,
    title = "Can a Suit of Armor Conduct Electricity? A New Dataset for Open Book Question Answering",
    author = "Mihaylov, Todor  and
      Clark, Peter  and
      Khot, Tushar  and
      Sabharwal, Ashish",
    editor = "Riloff, Ellen  and
      Chiang, David  and
      Hockenmaier, Julia  and
      Tsujii, Jun{'}ichi",
    booktitle = "Proceedings of the 2018 Conference on Empirical Methods in Natural Language Processing",
    month = oct # "-" # nov,
    year = "2018",
    address = "Brussels, Belgium",
    publisher = "Association for Computational Linguistics",
    url = "https://aclanthology.org/D18-1260/",
    doi = "10.18653/v1/D18-1260",
    pages = "2381--2391"
}

@misc{arc,
      title={Think you have Solved Question Answering? Try ARC, the AI2 Reasoning Challenge},
      author={Peter Clark and Isaac Cowhey and Oren Etzioni and Tushar Khot and Ashish Sabharwal and Carissa Schoenick and Oyvind Tafjord},
      year={2018},
      eprint={1803.05457},
      archivePrefix={arXiv},
      primaryClass={cs.AI},
      url={https://arxiv.org/abs/1803.05457},
}

@article{winogrande,
author = {Sakaguchi, Keisuke and Bras, Ronan Le and Bhagavatula, Chandra and Choi, Yejin},
title = {WinoGrande: an adversarial winograd schema challenge at scale},
year = {2021},
issue_date = {September 2021},
publisher = {Association for Computing Machinery},
address = {New York, NY, USA},
volume = {64},
number = {9},
issn = {0001-0782},
url = {https://doi.org/10.1145/3474381},
doi = {10.1145/3474381},
journal = {Commun. ACM},
month = aug,
pages = {99–106},
numpages = {8}
}

@inproceedings{hellaswag,
    title = "{H}ella{S}wag: Can a Machine Really Finish Your Sentence?",
    author = "Zellers, Rowan  and
      Holtzman, Ari  and
      Bisk, Yonatan  and
      Farhadi, Ali  and
      Choi, Yejin",
    editor = "Korhonen, Anna  and
      Traum, David  and
      M{\`a}rquez, Llu{\'i}s",
    booktitle = "Proceedings of the 57th Annual Meeting of the Association for Computational Linguistics",
    month = jul,
    year = "2019",
    address = "Florence, Italy",
    publisher = "Association for Computational Linguistics",
    url = "https://aclanthology.org/P19-1472/",
    doi = "10.18653/v1/P19-1472",
    pages = "4791--4800"
}

@inproceedings{piqa,
  title={{PIQA}: Reasoning about Physical Commonsense in Natural Language},
  author={Bisk, Yonatan and Zellers, Rowan and Le Bras, Ronan and Gao, Jianfeng and Choi, Yejin},
  booktitle={Proceedings of the AAAI Conference on Artificial Intelligence},
  volume={34},
  number={05},
  pages={7432--7439},
  year={2020},
  doi={10.1609/aaai.v34i05.6239}
}

@inproceedings{mathqa,
    title = "{M}ath{QA}: Towards Interpretable Math Word Problem Solving with Operation-Based Formalisms",
    author = "Amini, Aida  and
      Gabriel, Saadia  and
      Lin, Shanchuan  and
      Koncel-Kedziorski, Rik  and
      Choi, Yejin  and
      Hajishirzi, Hannaneh",
    editor = "Burstein, Jill  and
      Doran, Christy  and
      Solorio, Thamar",
    booktitle = "Proceedings of the 2019 Conference of the North {A}merican Chapter of the Association for Computational Linguistics: Human Language Technologies, Volume 1 (Long and Short Papers)",
    month = jun,
    year = "2019",
    address = "Minneapolis, Minnesota",
    publisher = "Association for Computational Linguistics",
    url = "https://aclanthology.org/N19-1245/",
    doi = "10.18653/v1/N19-1245",
    pages = "2357--2367"
}

@inproceedings{wangqsvd,
  title={QSVD: Efficient Low-rank Approximation for Unified Query-Key-Value Weight Compression in Low-Precision Vision-Language Models},
  author={Wang, Yutong and Wang, Haiyu and Zhang, Sai Qian},
  booktitle={The Thirty-ninth Annual Conference on Neural Information Processing Systems}
}

@inproceedings{wangwsvd,
  title={WSVD: Weighted Low-Rank Approximation for Fast and Efficient Execution of Low-Precision Vision-Language Models},
  author={Wang, Haiyu and Wang, Yutong and Jiang, Jack and Zhang, Sai Qian},
  booktitle={The Fourteenth International Conference on Learning Representations}
}

@inproceedings{lu2022learn_sqa,
    title={Learn to Explain: Multimodal Reasoning via Thought Chains for Science Question Answering},
    author={Lu, Pan and Mishra, Swaroop and Xia, Tony and Qiu, Liang and Chang, Kai-Wei and Zhu, Song-Chun and Tafjord, Oyvind and Clark, Peter and Ashwin Kalyan},
    booktitle={The 36th Conference on Neural Information Processing Systems (NeurIPS)},
    year={2022}
}

@inproceedings{li2024seed,
    title={Seed-bench: Benchmarking multimodal large language models},
    author={Li, Bohao and Ge, Yuying and Ge, Yixiao and Wang, Guangzhi and Wang, Rui and Zhang, Ruimao and Shan, Ying},
    booktitle={Proceedings of the IEEE/CVF Conference on Computer Vision and Pattern Recognition},
    pages={13299--13308},
    year={2024}
}

\clearpage

\appendix
\section*{Appendix}

\section{Adaptive-rank IO-SVD algorithm}
\label{sec:iosvd_algorithm}

Algorithm~\ref{alg:iosvd_truncation} summarizes the adaptive-rank IO-SVD truncation procedure. For each target layer, IO-SVD constructs the doubly whitened matrix
\[
B_\ell=C_\ell^{1/2}W_\ell R_\ell^{1/2},
\]
computes its SVD, and scores the first-order calibration-loss sensitivity of each whitened singular component. The algorithm maintains a shared pool containing one candidate from each layer: the smallest currently retained singular component of that layer. At each step, it removes the candidate with the smallest predicted loss impact while preserving the per-layer spectral order. This induces heterogeneous ranks across layers under a single global storage budget.

A practical detail is that low-rank factorization is storage-efficient only below a layer-dependent threshold. For a dense weight
\(W_\ell\in\mathbb{R}^{m_\ell\times n_\ell}\), the dense representation uses \(m_\ell n_\ell\) parameters, while a rank-\(r\) factorization uses \(r(m_\ell+n_\ell)\) parameters. We define
\begin{equation}
r_\ell^\star
:=
\left\lfloor
\frac{m_\ell n_\ell}{m_\ell+n_\ell}
\right\rfloor ,
\label{eq:threshold_rank}
\end{equation}
the largest rank at which the low-rank representation is no more expensive than the dense weight. Above this threshold, storing two low-rank factors would require more parameters than keeping the original dense module.

Let \(r\) denote the current rank before removing the tail singular component. The storage gain from dropping this component is
\begin{equation}
\label{eq:storage_gain}
\mathrm{storage\_gain}_\ell(r)=
\begin{cases}
0, & r>r_\ell^\star+1,\\[2pt]
m_\ell n_\ell-r_\ell^\star(m_\ell+n_\ell), & r=r_\ell^\star+1,\\[2pt]
m_\ell+n_\ell, & r\le r_\ell^\star .
\end{cases}
\end{equation}
When \(r>r_\ell^\star+1\), the layer remains cheaper to store densely even after the drop, so the realized storage gain is zero. When \(r=r_\ell^\star+1\), the next drop moves the layer to rank \(r_\ell^\star\), allowing the dense weight to be replaced by low-rank IO-SVD factors and yielding the one-time gain
\(m_\ell n_\ell-r_\ell^\star(m_\ell+n_\ell)\). Once the layer is already below the threshold, each additional rank reduction removes one column-row pair from the two factors and saves \(m_\ell+n_\ell\) parameters.

At the end of truncation, if a layer remains above the threshold rank, we keep the original dense module, i.e., \(\hat W_\ell\leftarrow W_\ell\). This avoids introducing approximation error when the low-rank representation would not provide storage savings.

\begin{algorithm}[b]
  \caption{IO-SVD initialization}
  \label{alg:iosvd_init}
  \begin{algorithmic}[1]
    \STATE {\bfseries Input:} weights \(\{W_\ell\}_{\ell=1}^L\), calibration set \(\mathcal{D}_{\mathrm{cal}}\), pruning ratio \(\pi\), minimum-rank ratio \(\eta\)
    \STATE {\bfseries Output:} per-layer state \(\mathcal{S}\), min-heap \(\mathcal{Q}\), target removed budget \(B_{\mathrm{rm}}\), removed budget \(b\)
    \STATE Set \(B_{\mathrm{rm}}\leftarrow \pi \sum_{\ell=1}^L m_\ell n_\ell\), \(b\leftarrow 0\), and initialize \(\mathcal{Q}\leftarrow\emptyset\)
    \FOR{\(\ell=1\) {\bfseries to} \(L\)}
      \STATE Estimate \(R_\ell\) and \(C_\ell\) on \(\mathcal{D}_{\mathrm{cal}}\)
      \STATE Form \(B_\ell\leftarrow C_\ell^{1/2}W_\ell R_\ell^{1/2}\) and compute \(B_\ell=U_\ell\Sigma_\ell V_\ell^\top\)
      \STATE Sort singular components in descending order of \(\sigma_{\ell,i}\), with corresponding columns of \(U_\ell\) and \(V_\ell\)
      \STATE Compute \(G_\ell=\partial\mathcal{L}_{\mathrm{cal}}/\partial W_\ell\) and \(\widetilde G_\ell\leftarrow C_\ell^{-1/2}G_\ell R_\ell^{-1/2}\)
      \STATE Set \(r_\ell\leftarrow r_\ell^{\max}=\min(m_\ell,n_\ell)\)
      \STATE Set \(r_\ell^\star\leftarrow \left\lfloor m_\ell n_\ell/(m_\ell+n_\ell)\right\rfloor\)
      \STATE Set \(r_\ell^{\min}\leftarrow \lceil \eta r_\ell^\star\rceil\)
      \FOR{\(i=1\) {\bfseries to} \(r_\ell^{\max}\)}
        \STATE \(g_{\ell,i}\leftarrow u_{\ell,i}^{\top}\widetilde G_\ell v_{\ell,i}\)
        \STATE \(I_{\ell,i}\leftarrow |g_{\ell,i}\sigma_{\ell,i}|\)
      \ENDFOR
      \IF{\(r_\ell>r_\ell^{\min}\)}
        \STATE \(\Delta b_{\ell,r_\ell}\leftarrow \mathrm{storage\_gain}_\ell(r_\ell)\)
        \STATE Push \((I_{\ell,r_\ell},\ell,r_\ell,\Delta b_{\ell,r_\ell})\) into \(\mathcal{Q}\)
      \ENDIF
    \ENDFOR
  \end{algorithmic}
\end{algorithm}

\begin{algorithm}[t]
  \caption{Adaptive-rank IO-SVD truncation}
  \label{alg:iosvd_truncation}
  \begin{algorithmic}[1]
    \STATE {\bfseries Input:} per-layer state \(\mathcal{S}\), min-heap \(\mathcal{Q}\), target removed budget \(B_{\mathrm{rm}}\), removed budget \(b\)
    \STATE {\bfseries Output:} compressed weights \(\{\hat W_\ell\}_{\ell=1}^L\)
    \WHILE{\(b<B_{\mathrm{rm}}\) {\bfseries and} \(\mathcal{Q}\neq\emptyset\)}
      \STATE Pop \((I_{\ell,i},\ell,i,\Delta b_{\ell,i})\) with the smallest score from \(\mathcal{Q}\)
      \STATE Drop the current tail singular component of layer \(\ell\)
      \STATE Update \(b\leftarrow b+\Delta b_{\ell,i}\) and \(r_\ell\leftarrow r_\ell-1\)
      \IF{\(r_\ell>r_\ell^{\min}\)}
        \STATE \(\Delta b_{\ell,r_\ell}\leftarrow \mathrm{storage\_gain}_\ell(r_\ell)\)
        \STATE Push \((I_{\ell,r_\ell},\ell,r_\ell,\Delta b_{\ell,r_\ell})\) into \(\mathcal{Q}\)
      \ENDIF
    \ENDWHILE

    \FOR{\(\ell=1\) {\bfseries to} \(L\)}
      \IF{\(r_\ell>r_\ell^\star\)}
        \STATE Keep dense module: \(\hat W_\ell\leftarrow W_\ell\)
      \ELSE
        \STATE \(\hat B_\ell\leftarrow U_{\ell,1:r_\ell}\Sigma_{\ell,1:r_\ell}V_{\ell,1:r_\ell}^{\top}\)
        \STATE \(\hat W_\ell\leftarrow C_\ell^{-1/2}\hat B_\ell R_\ell^{-1/2}\)
      \ENDIF
    \ENDFOR
  \end{algorithmic}
\end{algorithm}

\section{Objective derivation}
\label{app:objective_derivation}

\subsection{KL expansion and output-side curvature}
\label{app:kl_whitening}

We derive the second-order approximation used in Eq.~\eqref{eq:kl_second_order}. For a fixed token \(t\), let
\[
p_t=\mathrm{softmax}(z_t)
\]
be the predictive distribution of the uncompressed model, and define
\begin{equation}
f_t(z)
:=
\mathrm{KL}\!\left(p_t\,\|\,\mathrm{softmax}(z)\right).
\end{equation}
Let \(q(z)=\mathrm{softmax}(z)\). Then
\begin{equation}
f_t(z)
=
\sum_{i=1}^V p_{t,i}\log p_{t,i}
-
\sum_{i=1}^V p_{t,i}\log q_i(z).
\end{equation}
Using
\[
\log q_i(z)=z_i-\log\sum_{j=1}^V e^{z_j},
\]
we obtain
\begin{equation}
f_t(z)
=
\sum_{i=1}^V p_{t,i}\log p_{t,i}
-
p_t^\top z
+
\log\sum_{j=1}^V e^{z_j}.
\end{equation}
Therefore,
\begin{equation}
\nabla_z f_t(z)=q(z)-p_t.
\end{equation}
At \(z=z_t\), \(q(z_t)=p_t\), so
\begin{equation}
\nabla_z f_t(z_t)=0.
\end{equation}
Since \(f_t(z_t)=\mathrm{KL}(p_t\,\|\,p_t)=0\), both the zeroth- and first-order terms vanish.

The Hessian is the softmax Jacobian:
\begin{equation}
\nabla_z^2 f_t(z)
=
\mathrm{Diag}(q(z))-q(z)q(z)^\top .
\end{equation}
Evaluating at \(z=z_t\) gives
\begin{equation}
H_t
=
\nabla_z^2 f_t(z_t)
=
\mathrm{Diag}(p_t)-p_tp_t^\top .
\label{eq:app_hessian}
\end{equation}
Thus, for a logit perturbation \(\delta z_t\),
\begin{equation}
\mathrm{KL}\!\left(p_t\,\|\,\mathrm{softmax}(z_t+\delta z_t)\right)
=
\frac{1}{2}\delta z_t^\top H_t\delta z_t
+
O(\|\delta z_t\|^3).
\end{equation}

The matrix \(H_t\) is symmetric positive semidefinite. For any \(v\in\mathbb{R}^V\),
\begin{equation}
v^\top H_t v
=
\sum_{i=1}^V p_{t,i}v_i^2
-
\left(\sum_{i=1}^V p_{t,i}v_i\right)^2
\ge 0,
\end{equation}
which is the variance of \(v_i\) under \(p_t\). Also, \(H_t\mathbf{1}=0\), reflecting the invariance of the softmax distribution to adding a constant to all logits.

\subsection{Trace-to-Frobenius reduction}
\label{app:trace_to_frobenius}

Starting from Eq.~\eqref{eq:token_local_surrogate},
\begin{equation}
\Delta\mathcal{J}_{\ell,t}
\approx
\frac{1}{2}
x_t^\top
\Delta W_\ell^\top
C_{\mathrm{token},t}
\Delta W_\ell
x_t ,
\end{equation}
we use \(a^\top M a=\mathrm{tr}(Maa^\top)\) to write
\begin{equation}
x_t^\top
\Delta W_\ell^\top
C_{\mathrm{token},t}
\Delta W_\ell
x_t
=
\mathrm{tr}
\left(
\Delta W_\ell x_tx_t^\top \Delta W_\ell^\top C_{\mathrm{token},t}
\right).
\end{equation}
Averaging over calibration tokens gives
\begin{equation}
\Delta\mathcal{J}_\ell
\approx
\frac{1}{2}
\mathbb{E}_t
\left[
\mathrm{tr}
\left(
\Delta W_\ell x_tx_t^\top \Delta W_\ell^\top C_{\mathrm{token},t}
\right)
\right].
\end{equation}
Using the moment-decoupling approximation,
\begin{equation}
R_\ell=\mathbb{E}_t[x_tx_t^\top],
\qquad
C_\ell=\mathbb{E}_t[C_{\mathrm{token},t}],
\end{equation}
we obtain
\begin{equation}
\Delta\mathcal{J}_\ell
\approx
\frac{1}{2}
\mathrm{tr}
\left(
\Delta W_\ell R_\ell \Delta W_\ell^\top C_\ell
\right).
\end{equation}
Let
\[
M=C_\ell^{1/2}\Delta W_\ell R_\ell^{1/2}.
\]
Then
\begin{align}
\|M\|_F^2
&=
\mathrm{tr}(MM^\top) \\
&=
\mathrm{tr}
\left(
C_\ell^{1/2}\Delta W_\ell R_\ell
\Delta W_\ell^\top C_\ell^{1/2}
\right) \\
&=
\mathrm{tr}
\left(
\Delta W_\ell R_\ell
\Delta W_\ell^\top C_\ell
\right),
\end{align}
where the last equality uses cyclic invariance of trace. Therefore,
\begin{equation}
\Delta\mathcal{J}_\ell
\approx
\frac{1}{2}
\left\|
C_\ell^{1/2}
\Delta W_\ell
R_\ell^{1/2}
\right\|_F^2.
\end{equation}
Substituting \(\Delta W_\ell=W_\ell-\hat W_\ell\) yields Eq.~\eqref{eq:double_whitened_objective}.

\section{Efficient computation of \texorpdfstring{\(J^\top HJ\)}{JTHJ}}
\label{app:efficient_curvature}

For a target module \(\ell\), let \(h_{\ell,t}\in\mathbb{R}^{d_{\mathrm{out}}}\) denote its output at token \(t\). Let \(z_{t,K}\in\mathbb{R}^{K}\) denote the logits restricted to the top-\(K\) coordinates of the uncompressed model, and let \(p_{t,K}\in\mathbb{R}^{K}\) be the corresponding probabilities renormalized on this support. The reduced token-level curvature is
\begin{equation}
C_{\mathrm{token},t}
\approx
\left(J_{t,K}^{(\ell)}\right)^\top
H_{t,K}
J_{t,K}^{(\ell)},
\qquad
J_{t,K}^{(\ell)}
=
\frac{\partial z_{t,K}}{\partial h_{\ell,t}},
\end{equation}
where
\begin{equation}
H_{t,K}
=
\mathrm{Diag}(p_{t,K})-p_{t,K}p_{t,K}^\top .
\end{equation}

Directly forming \(J_{t,K}^{(\ell)}\) for all tokens and modules is expensive. Instead, we exploit the structure of \(H_{t,K}\). Define
\begin{equation}
s_t=\sqrt{p_{t,K}},
\qquad
D_t=\mathrm{Diag}(s_t),
\qquad
\Omega_t=I-s_ts_t^\top .
\end{equation}
Since \(p_{t,K}\) is renormalized, \(\|s_t\|_2=1\), and hence
\begin{equation}
\Omega_t^2
=
(I-s_ts_t^\top)^2
=
I-s_ts_t^\top
=
\Omega_t .
\end{equation}
Thus, \(\Omega_t\) is an orthogonal projector onto the subspace orthogonal to \(s_t\).

Now define
\begin{equation}
A_t=D_t\Omega_t .
\end{equation}
Then
\begin{align}
A_tA_t^\top
&=
D_t\Omega_t^2D_t \\
&=
D_t\Omega_t D_t \\
&=
D_t(I-s_ts_t^\top)D_t \\
&=
\mathrm{Diag}(p_{t,K})-p_{t,K}p_{t,K}^\top
=
H_{t,K}.
\end{align}
Therefore,
\begin{equation}
H_{t,K}=A_tA_t^\top .
\label{eq:app_topk_factorization}
\end{equation}

Let \(e_j\) be the \(j\)-th canonical basis vector in \(\mathbb{R}^{K}\), and define the deterministic probe
\begin{equation}
v_{t,j}=A_t e_j .
\end{equation}
Sweeping over all \(K\) basis directions recovers the reduced Hessian:
\begin{equation}
\sum_{j=1}^{K}v_{t,j}v_{t,j}^\top
=
A_t
\left(
\sum_{j=1}^{K}e_je_j^\top
\right)
A_t^\top
=
A_tA_t^\top
=
H_{t,K}.
\end{equation}

For each probe \(v_{t,j}\), define the scalar
\begin{equation}
\phi_{t,j}=v_{t,j}^\top z_{t,K}.
\end{equation}
Its gradient with respect to the module output is
\begin{equation}
\nabla_{h_{\ell,t}}\phi_{t,j}
=
\left(J_{t,K}^{(\ell)}\right)^\top v_{t,j}.
\end{equation}
This is a vector-Jacobian product and can be computed by reverse-mode automatic differentiation without materializing \(J_{t,K}^{(\ell)}\). Denote
\begin{equation}
g_{\ell,t,j}
=
\nabla_{h_{\ell,t}}\phi_{t,j}
=
\left(J_{t,K}^{(\ell)}\right)^\top v_{t,j}.
\end{equation}
Then
\begin{align}
\sum_{j=1}^{K}g_{\ell,t,j}g_{\ell,t,j}^\top
&=
\left(J_{t,K}^{(\ell)}\right)^\top
\left(
\sum_{j=1}^{K}v_{t,j}v_{t,j}^\top
\right)
J_{t,K}^{(\ell)} \\
&=
\left(J_{t,K}^{(\ell)}\right)^\top
H_{t,K}
J_{t,K}^{(\ell)}.
\end{align}
Thus, the curvature is recovered exactly on the top-\(K\) subspace; the only approximation is the restriction from the full vocabulary to the top-\(K\) support.

In practice, for each calibration batch, we compute the top-\(K\) logits and renormalized probabilities, form the probes \(\{v_{t,j}\}_{j=1}^{K}\), and backpropagate each scalar \(\phi_{t,j}\). Backward hooks on each target linear module collect the vectors \(g_{\ell,t,j}\) and accumulate
\begin{equation}
\widehat C_\ell
=
\frac{1}{N}
\sum_{t=1}^{N}
\sum_{j=1}^{K}
g_{\ell,t,j}g_{\ell,t,j}^\top
\approx
\mathbb{E}_t
\left[
J_t^\top H_tJ_t
\right].
\end{equation}
A small damping term may be added to \(\widehat C_\ell\) before taking matrix square roots or inverse square roots.

\section{Derivation of the singular-component score}
\label{app:rank_score_derivation}

The adaptive rank-allocation score in Eq.~\eqref{eq:singular_component_score} follows from a first-order Taylor approximation in the whitened SVD coordinates. Recall that
\begin{equation}
B_\ell=C_\ell^{1/2}W_\ell R_\ell^{1/2}.
\end{equation}
A perturbation \(\Delta W_\ell\) induces
\begin{equation}
\Delta B_\ell
=
C_\ell^{1/2}\Delta W_\ell R_\ell^{1/2},
\qquad
\Delta W_\ell
=
C_\ell^{-1/2}\Delta B_\ell R_\ell^{-1/2}.
\end{equation}
Let \(G_\ell=\partial\mathcal{L}/\partial W_\ell\). The first-order loss change is
\begin{align}
\Delta\mathcal{L}
&\approx
\langle G_\ell,\Delta W_\ell\rangle \\
&=
\mathrm{tr}
\left(
G_\ell^\top
C_\ell^{-1/2}
\Delta B_\ell
R_\ell^{-1/2}
\right) \\
&=
\mathrm{tr}
\left(
\left(C_\ell^{-1/2}G_\ell R_\ell^{-1/2}\right)^\top
\Delta B_\ell
\right) \\
&=
\langle \widetilde G_\ell,\Delta B_\ell\rangle,
\end{align}
where
\begin{equation}
\widetilde G_\ell
=
C_\ell^{-1/2}G_\ell R_\ell^{-1/2}
=
\frac{\partial\mathcal{L}}{\partial B_\ell}.
\end{equation}

Let
\begin{equation}
B_\ell
=
\sum_i
\sigma_{\ell,i}u_{\ell,i}v_{\ell,i}^\top .
\end{equation}
Holding \(u_{\ell,i}\) and \(v_{\ell,i}\) fixed, we have
\begin{equation}
\frac{\partial B_\ell}{\partial\sigma_{\ell,i}}
=
u_{\ell,i}v_{\ell,i}^\top .
\end{equation}
Therefore,
\begin{equation}
g_{\ell,i}
=
\frac{\partial\mathcal{L}}{\partial\sigma_{\ell,i}}
=
\left\langle
\widetilde G_\ell,
u_{\ell,i}v_{\ell,i}^\top
\right\rangle
=
u_{\ell,i}^\top
\widetilde G_\ell
v_{\ell,i}.
\end{equation}
Removing the \(i\)-th singular component sets
\[
\Delta\sigma_{\ell,i}=-\sigma_{\ell,i}.
\]
The predicted first-order loss change is
\begin{equation}
\Delta\mathcal{L}_{\ell,i}
\approx
g_{\ell,i}\Delta\sigma_{\ell,i}
=
-g_{\ell,i}\sigma_{\ell,i}.
\end{equation}
We use its magnitude as the component importance score:
\begin{equation}
I_{\ell,i}
=
|g_{\ell,i}\sigma_{\ell,i}|.
\end{equation}

\section{Additional details for loss-aware remapping}
\label{app:remapping_details}

For hybrid SVD--quantization compression, we write each truncated module as
\begin{equation}
\hat W_\ell=A_\ell D_\ell^\top,
\end{equation}
where \(A_\ell\in\mathbb{R}^{d_{\mathrm{out}}\times r_\ell}\) and
\(D_\ell\in\mathbb{R}^{d_{\mathrm{in}}\times r_\ell}\). Candidate rows are drawn from both low-rank factors. For a candidate row \(r_{\ell,i}\), the 8-bit quantize-dequantize perturbation is
\begin{equation}
\Delta r_{\ell,i}
=
\mathcal{Q}_8(r_{\ell,i})-r_{\ell,i}.
\end{equation}
Let
\begin{equation}
\gamma_{\ell,i}
=
\frac{\partial\mathcal{L}_{\mathrm{cal}}}{\partial r_{\ell,i}}
\end{equation}
be the row gradient computed on calibration data after SVD truncation. A first-order Taylor approximation gives
\begin{equation}
\Delta\mathcal{L}_{\ell,i}
\approx
\langle \gamma_{\ell,i},\Delta r_{\ell,i}\rangle
=
\left\langle
\gamma_{\ell,i},
\mathcal{Q}_8(r_{\ell,i})-r_{\ell,i}
\right\rangle .
\end{equation}
We define the remapping sensitivity score as
\begin{equation}
s_{\ell,i}
=
\left|
\left\langle
\gamma_{\ell,i},
\mathcal{Q}_8(r_{\ell,i})-r_{\ell,i}
\right\rangle
\right|.
\end{equation}
This score is small when either the row is quantized accurately or the quantization error lies in a direction that has little effect on the calibration loss.

Let \(C_{\mathrm{svd}}\) denote the storage saved by SVD truncation and \(C_{\mathrm{target}}\) denote the desired total storage saving. The remaining saving required from quantization is
\begin{equation}
C_{\mathrm{rem}}
=
\max\{0, C_{\mathrm{target}}-C_{\mathrm{svd}}\}.
\end{equation}
Each candidate row has a storage saving \(c_{\ell,i}\), which may depend on the row length and quantization format. We select rows by solving the budgeted selection problem
\begin{equation}
\min_{\mathcal{S}}
\sum_{(\ell,i)\in\mathcal{S}} s_{\ell,i}
\quad
\text{subject to}
\quad
\sum_{(\ell,i)\in\mathcal{S}} c_{\ell,i}
\ge
C_{\mathrm{rem}}.
\end{equation}
In practice, we use a greedy approximation. When all candidate rows have the same storage saving, rows are selected in increasing order of \(s_{\ell,i}\). When row sizes differ, rows are selected by increasing cost per saved parameter:
\begin{equation}
\frac{s_{\ell,i}}{c_{\ell,i}}.
\end{equation}
The selected rows are stored in 8-bit precision, and their row indices are stored as metadata. This index overhead is negligible relative to the compressed model size.


\end{document}